\documentclass[runningheads]{llncs}

\usepackage{eccv}

\usepackage[accsupp]{axessibility}  

\usepackage{eccvabbrv}

\usepackage{graphicx}
\usepackage{booktabs}
\usepackage{subcaption}
\usepackage{multirow}
\usepackage{amssymb}
\usepackage{duckuments}
\usepackage{caption}
\usepackage[table]{xcolor}
\usepackage{adjustbox}
\usepackage[most]{tcolorbox}
\usepackage{stackengine}
\usepackage{pifont}

\usepackage[ruled,vlined]{algorithm2e}
\tcbuselibrary{breakable,listings}
\usepackage{wrapfig}
\usepackage{paralist}
\usepackage{lipsum}
\usepackage{makecell}
\usepackage{boldline}
\usepackage{mdframed}
\usepackage{lmodern}
\usepackage{textcomp}
\usepackage{listings}
\usepackage{siunitx}

\lstdefinestyle{logstyle}{
  basicstyle=\ttfamily\footnotesize,
  breaklines=true,
  columns=fullflexible,
  showstringspaces=false,
  keepspaces=true,
  frame=none
}
\lstdefinestyle{jsonstyle}{
  basicstyle=\ttfamily\footnotesize,
  breaklines=true,
  columns=fullflexible,
  showstringspaces=false,
  keepspaces=true,
  frame=none
}
\newmdenv[
  linewidth=0.5pt,
  roundcorner=2pt,
  innerleftmargin=8pt,
  innerrightmargin=8pt,
  innertopmargin=6pt,
  innerbottommargin=6pt,
  skipabove=4pt,
  skipbelow=0pt
]{codebox}

\graphicspath{{assert/fig/}}
\newcommand{\ours}{MERIT}
\renewcommand{\subsubsection}[1]{\vspace{1pt}\noindent\textbf{#1}~}

\usepackage{hyperref}

\usepackage{orcidlink}

\begin{document}

\title{Keep It Simple: Multi-Key Episodic Memory Retrieval for Ultra-Long Video Understanding}

\titlerunning{MERIT}

\author{Yeeun Choi$^{1}$ \and
Youngbeom Yoo$^{1}$ \and
Joon-Young Lee$^{2}$ \and \\
Hyolim Kang$^{1,\dagger}$ \and
Seon Joo Kim$^{1,\dagger}$}

\authorrunning{Y.~Choi et al.}

\institute{$^{1}$Yonsei University \quad $^{2}$Adobe Research}

\maketitle

\begin{center}
\vspace{-3.5mm}
  \textbf{Project Page:}~\href{https://choi-yeeun.github.io/MERIT}{\url{https://choi-yeeun.github.io/MERIT}}
\end{center}
\vspace{-6.5mm}

\let\thefootnote\relax\footnotetext{\textsuperscript{$\dagger$}Co-corresponding authors.}

\begin{abstract}
When videos extend from hours to days, directly processing them end-to-end becomes impractical for current Multi-modal Large Language Models (MLLMs). This ultra-long setting necessitates a two-stage paradigm: query-agnostic memory construction followed by retrieval-based inference. Prior work invests in complex memory construction to pre-model high-level relations in videos, despite not knowing the downstream query at build time. We instead prioritize high-recall retrievability during memory building, and defer query-specific, high-level relation composition to inference time.
To this end, we propose \textbf{\ours{}} (\textbf{M}ulti-key \textbf{E}pisodic \textbf{R}etrieval with \textbf{I}nference-time \textbf{T}emporal expansion), a simple yet effective agentic framework for ultra-long video understanding. First, we formulate an episodic multi-key representation that enables precise retrieval of fine-grained memories through a simple key-matching mechanism. Second, we introduce a neighbor filtering mechanism to capture broader semantic context without the massive computational overhead of global memory construction. This is achieved by expanding the temporal scope exclusively around the retrieved segments at inference time. By leveraging simple key-matching with this on-demand temporal expansion, \ours{} achieves state-of-the-art performance across three long-video benchmarks: EgoLifeQA, LVBench, and Video-MME (Long).

\end{abstract}
\section{Introduction}
\label{sec:intro}

The rapid evolution of Multi-modal Large Language Models (MLLMs)~\cite{gpt5, gemini2.5, qwen3-vl, qwen2.5-vl, llava-video} has shifted video understanding from short temporal clips to ultra-long streams spanning hours or even days. This shift enables the development of personal AI assistants capable of agentic reasoning over continuous egocentric or surveillance video. However, as temporal length scales dramatically, direct end-to-end modeling becomes infeasible, making external memory and Retrieval-Augmented Generation (RAG)~\cite{rag, gao2023ragsurvey, edge2024local} frameworks essential.

A defining property of ultra-long video understanding is the structural separation between memory construction time and inference time. During memory construction, the system processes the entire video without knowledge of future queries and builds an external memory. At inference time, a specific question is posed, and the system must retrieve relevant evidence and perform reasoning. This decoupling fundamentally distinguishes long-video settings from conventional short-video Question Answering (QA).

\begin{figure}[t]
  \centering
  \setlength{\abovecaptionskip}{4pt}
  \setlength{\belowcaptionskip}{2pt}
  \includegraphics[width=1\linewidth]{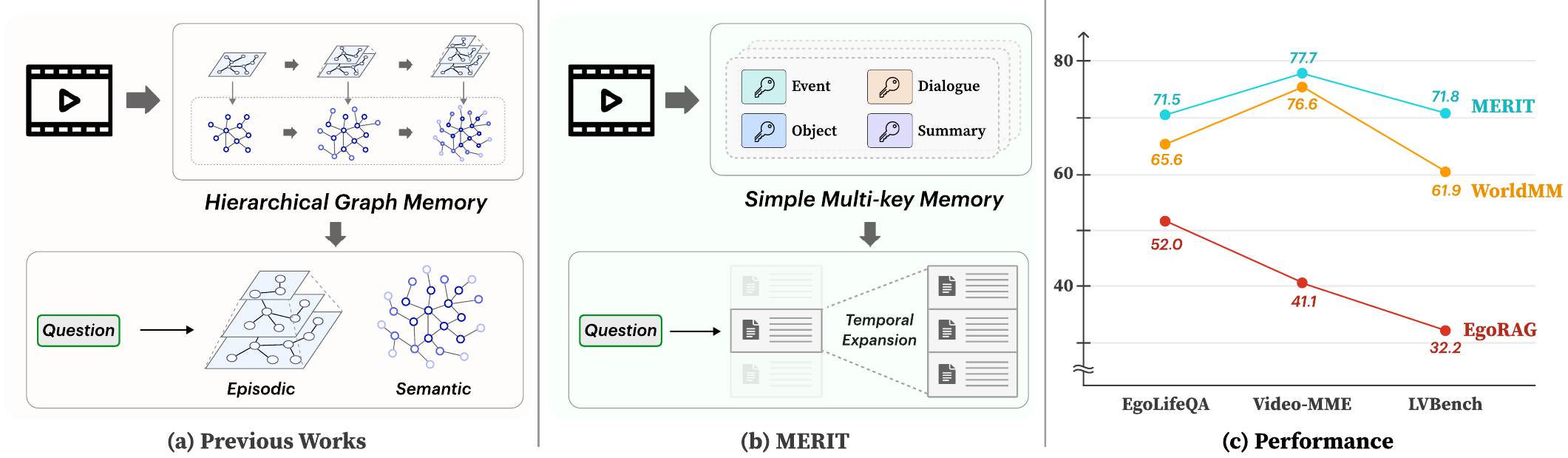}
  \caption{(a) Previous methods: Construct hierarchical graph-based episodic and semantic memories, and at inference time, search and aggregate relevant information across these structured memories. (b) Our method: \ours{} constructs a simple episodic multi-key memory, and at inference time, retrieves the matched clips along with temporal expansion to perform query-driven semantic reasoning. (c) Performance of \ours{} against EgoRAG~\cite{egolife} and WorldMM~\cite{worldmm} across various benchmarks~\cite{egolife, videomme, lvbench}.}
  \label{fig:fig1}
\end{figure}

To address long-range dependencies, prior work has proposed increasingly sophisticated memory architectures. Hierarchical frameworks~\cite{egolife, egor1} pre-compute multi-scale temporal representations to enable structured retrieval, while graph-based methods~\cite{vgent,worldmm} explicitly model relational dependencies across events. These designs are motivated by the assumption that naive segment-level retrieval is insufficient for capturing high-level semantics (Fig.~\ref{fig:fig1}(a)).

However, constructing structured memories is not only computationally heavy but also reasoning-demanding. For instance, EgoRAG~\cite{egolife} builds a hierarchical memory by progressively merging 30-second segments into hour-level and day-level summaries, requiring multiple rounds of MLLM inference at each temporal scale. WorldMM~\cite{worldmm} goes further: it constructs multi-scale knowledge graphs across different temporal resolutions, where each scale requires caption generation, triplet extraction, and graph consolidation via a Large Language Model (LLM) that continuously merges new knowledge while resolving conflicts with existing relations. As video length increases, these multi-stage processes grow rapidly in both computational cost and MLLM calls. More importantly, these structuring decisions are made without access to the downstream query, forcing a query-agnostic abstraction that may not align with what is ultimately asked.

This reveals a practical issue of \emph{Intelligence Allocation}. If strong MLLMs are available, using them primarily for query-agnostic preprocessing is inefficient: it expends high-capability reasoning before the task is even specified. A more principled alternative is to preserve high-fidelity episodic evidence and defer semantic composition to inference time, when the query determines which relations and abstractions matter. From this perspective, the bottleneck shifts from building static semantics in advance to enabling high-recall retrieval that consistently delivers the right evidence.

Motivated by this view, we adopt a simpler design (Fig.~\ref{fig:fig1}(b)): rather than constructing complex semantic hierarchies during memory building, we retrieve over minimally processed episodic segments and form task-specific relations on demand. To this end, we introduce \ours, a lightweight yet effective framework built on two key ideas: multi-key indexing and neighbor filtering.

First, we propose a \textbf{Multi-key Indexing} strategy. Rather than constructing hierarchical or graph-based structures, we attach multiple complementary keys to each minimal temporal segment. These keys capture different perspectives—such as event-centric, object-centric, dialogue-centric, or summary-level cues—allowing diverse queries to match relevant evidence without requiring explicit multi-scale memory design. By enriching each fine-grained segment with heterogeneous semantic anchors, we substantially improve retrieval robustness while keeping memory construction simple.

Second, we introduce \textbf{Neighbor Filtering}, a query-aware local aggregation mechanism grounded in the inductive bias of temporal continuity. When a segment is retrieved via key matching, its temporally adjacent neighbors are jointly considered to form a local evidence cluster. Within this cluster, we perform query-conditioned relevance selection to extract information most pertinent to the query. In other words, neighbor filtering consists of both local temporal expansion and query-aware evidence refinement, enabling coherent context reconstruction without pre-computed hierarchical structures.

Through these two simple components, our framework eliminates the need for expensive multi-stage memory consolidation while maintaining strong retrieval quality. As shown in Fig.~\ref{fig:fig1}(c), \ours{} achieves state-of-the-art performance across multiple long-video QA benchmarks, including EgoLifeQA\cite{egolife}, LVBench~\cite{lvbench} and Video-MME\cite{videomme}, outperforming prior methods such as EgoRAG\cite{egolife} and WorldMM\cite{worldmm} with a significantly simpler memory pipeline. These results validate our central hypothesis: deferring semantic composition to query time, rather than investing in query-agnostic preprocessing, yields both better performance and greater efficiency. 
\section{Related Work}

\subsection{Ultra-long Video Understanding with MLLMs}

The paradigm of video understanding has shifted from short-clip analysis to extended temporal reasoning. While recent proprietary MLLMs~\cite{gpt5, gemini2.5} and open-source models~\cite{longvu, qwen3-vl, llava-video, zhang2024llavanext-video, li2024llavaonevision, chen2024internvl2_5, zhang2025videollama} can natively process hour-long videos via extended context windows, the emergence of ultra-long benchmarks~\cite{egolife, egor1} has pushed context requirements beyond these capacities. Despite increased window sizes, processing such extreme scales remains computationally prohibitive. Standard MLLMs under strict memory limits often resort to sparse uniform sampling, which inevitably discards fine-grained details and misses critical events. 

\subsubsection{Our Work.}
Rather than modifying or retraining the MLLM backbone, \ours{} leverages off-the-shelf MLLMs by providing them with only query-relevant evidence. Since these models are highly capable yet constrained by limited context windows, the key is to feed them the evidence that matters for each query rather than the entire video. To this end, \ours{} avoids heavy query-agnostic memory construction and instead maintains a lightweight memory for inference time retrieval and evidence curation. This defers query-specific reasoning to inference, where the model’s capacity can be used most effectively.

\subsection{Memory-based Architectures for Video QA}

To overcome the context window limitations of MLLMs, recent studies~\cite{egolife, egor1, hipporag, video-rag,videorag,evrag,ren2025videorag, adavideorag, vgent,hippomm,m3-agent,worldmm} have adopted the Retrieval-Augmented Generation (RAG) paradigm for the video domain. These frameworks construct a structured external memory from video data, enabling the system to retrieve and incorporate relevant visual evidence at inference time.

\subsubsection{Hierarchy-based Memory.}
Temporal hierarchy is a widely adopted structure for managing long-form video memory. EgoRAG~\cite{egolife} organizes video data into multiple temporal scales by recursively summarizing short 30-second segments. While this multi-level indexing provides a broad overview, the recursive summarization process inevitably discards fine-grained details. Furthermore, such top-down retrieval is highly sensitive to initial errors; a mismatch at the coarse summary level often leads to total QA failure. To mitigate these structural weaknesses, Ego-R1~\cite{egor1} introduces an agentic approach that utilizes multi-turn tool calling to dynamically navigate the hierarchical memory. However, performance remains fundamentally limited by the resolution of the underlying summaries.

\subsubsection{Graph-based Memory.}
Another prominent direction involves representing video memory as a graph to capture complex relational context. Techniques from text-based RAG~\cite{lightrag, hipporag} have been extended to the multimodal domain~\cite{videorag}, incorporating visual features into structured knowledge graphs. Models such as HippoRAG~\cite{hipporag} and HippoMM~\cite{hippomm} utilize short-to-long-term memory consolidation to build semantic graphs, while M3-Agent~\cite{m3-agent} incorporates entity-centric episodic and semantic memory. WorldMM~\cite{worldmm} further utilizes multiple graph-based memories for adaptive retrieval, and EGAgent~\cite{egagent} constructs time-aware graphs to improve temporal reasoning. 
However, graph-based approaches require intensive pre-computation for graph construction, incurring significant computational overhead and latency. For instance, WorldMM~\cite{worldmm} not only constructs multi-granular episodic graphs but also aggregates them into a global semantic graph, which inevitably necessitates continuous memory consolidation as the representation expands. Consequently, managing such cascading structural updates becomes prohibitively expensive as video length scales to multiple days.

\subsubsection{Our Work.}
In contrast to pre-structured hierarchies or graphs, \ours{} utilizes a multi-key episodic memory that bypasses the intensive preprocessing required for complex memory construction. By maintaining simple yet effective keys, we preserve fine-grained episodic details while minimizing memory building costs. Our framework achieves both efficiency and semantic depth by performing on-demand temporal expansion during inference, providing a scalable solution for ultra-long video understanding.

\subsection{Caption-based Video RAG}

An alternative paradigm represents video understanding in the language space, converting visual content into captions and delegating reasoning to a powerful LLM. LLoVi~\cite{llovi} densely captions short clips and aggregates them with an LLM, SiLVR~\cite{silvr} extends this caption-and-reason recipe with multisensory descriptions fed into a dedicated reasoning LLM, and VideoTree~\cite{videotree} organizes frames into a query-adaptive hierarchical tree for coarse-to-fine reasoning. As videos grow longer, video RAG systems build on such textual representations and retrieve only the query-relevant evidence from them at inference time~\cite{video-rag, adavideorag, goldfish}. Since retrieval is performed after the query is given, several works further generate query-aware, more informative evidence at this stage. For instance, iRAG~\cite{irag} keeps only a lightweight index after preprocessing. For each query, it runs heavier vision models on the retrieved clips to caption details that were not extracted beforehand. DrVideo~\cite{drvideo}, instead, starts from a coarse document and runs an agentic loop, re-captioning key frames until the gathered information suffices. 
 
\subsubsection{Our Work.}
\ours{} is likewise a caption-based, agentic video RAG framework. To obtain query-aware evidence, however, \ours{} does not re-run captioning on the raw video at every inference step. Instead, it retrieves diverse relevant clips through complementary multi-keys enabling high-recall retrieval, and then distills query-aware, rich information from the pre-built captions via neighbor filtering. In doing so, \ours{} also reconstructs the surrounding temporal context that single-clip retrieval would otherwise miss, without re-captioning per query.
\section{Method}

In this section, we present \ours{}, a minimalist agentic framework designed to improve retrieval accuracy through a straightforward and simplified memory structure for ultra-long video QA. Unlike prior approaches~\cite{worldmm} that rely on complex hierarchical graph construction, our method maintains minimal episodic representations and performs query-driven temporal expansion on demand.

\subsection{Problem Formulation}

In the ultra-long video setting, directly performing question answering (QA) by feeding raw video frames into MLLMs is infeasible due to context length limitations. Therefore, the common paradigm constructs an external memory representation $M$ from the video $V$: 
\begin{equation}
M=\mathcal{F}(V),    
\end{equation}
where $\mathcal{F}(\cdot)$ denotes the memory construction procedure.
The memory is built in a query-agnostic manner, meaning that the natural language query $Q$ is not available during memory construction. 
At inference time, the QA solver takes the query $Q$ and the pre-built memory $M$ to produce the final answer $A$:
\begin{equation}
A=\mathcal{G}(Q, M),    
\end{equation}
where $\mathcal{G}(\cdot)$ denotes the QA solver.

\begin{figure}[t]
  \centering
  \setlength{\abovecaptionskip}{4pt}
  \setlength{\belowcaptionskip}{2pt}
  \includegraphics[width=1\linewidth]
  {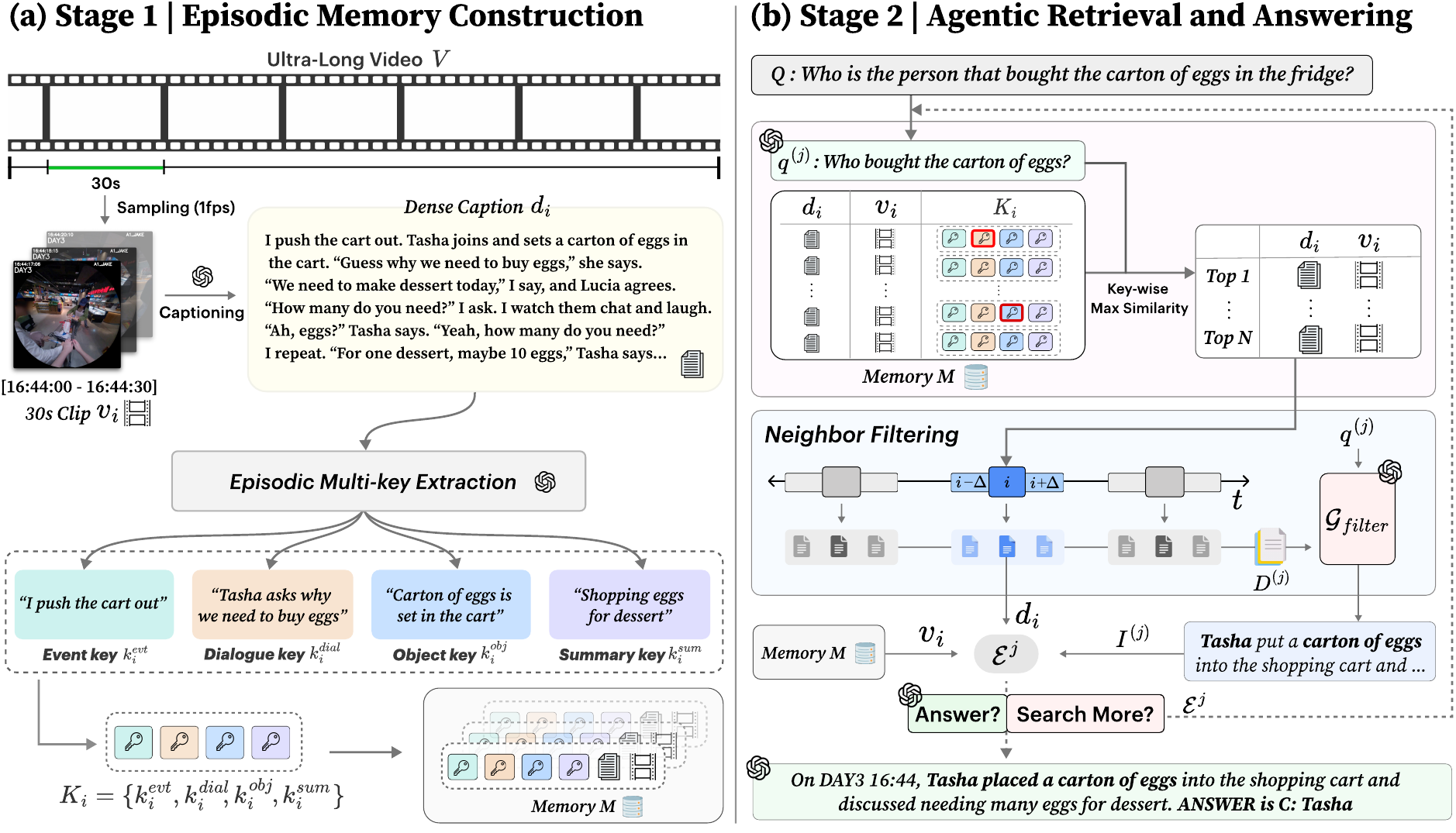}
  \caption{Overall Pipeline of \ours{}. (a) \textbf{Stage 1: Episodic memory construction.} For each 30-second video clip, dense captions are generated. Subsequently, an episodic multi-key extraction process derives four distinct keys per clip, collectively forming the memory $M$. (b) \textbf{Stage 2: Agentic retrieval and answering.} The solver regenerates a query and matches it against multi-keys to retrieve the most relevant clips. Neighbor Filtering then expands the temporal context around these clips, extracting additional query-relevant information to formulate the final answer.}
  \label{fig:fig2}
\end{figure}

\subsection{Overall Pipeline}

Fig.~\ref{fig:fig2} illustrates the overall pipeline of our method \ours{}. \ours{} follows a minimalist memory-based agentic pipeline, with an emphasis on high-recall retrieval and lightweight preprocessing.

\subsubsection{Episodic memory construction.}
We partition the video $V$ into a sequence of non-overlapping clips: $V = \{v_1, v_2, \dots, v_T\}$. We generate a dense caption $d_i$ for each clip $v_i$. For each clip, we additionally derive a set of textual keys $K_i$. The resulting memory is:
\begin{equation}
\label{eq:memory_store}
M = \mathcal{F}(V) = \{(v_i, d_i, K_i)\}_{i=1}^{T}.
\end{equation}
Conceptually, $M$ can be viewed as a minimal key--value store, where $K_i$ serves as retrieval keys and $(v_i, d_i)$ provides the associated values.

\subsubsection{Agentic retrieval and answering.}
During QA, the solver $\mathcal{G}$ runs an iterative multi-turn procedure~\cite{egor1, worldmm, m3-agent}. 
At round $j$, the solver forms a retrieval query $q^{(j)}$ from $Q$ and the retrieval results accumulated up to the previous round, namely \emph{evidence} $\mathcal{E}^{(j-1)}$, with $\mathcal{E}^{(0)}=\emptyset$.
It then retrieves a set of relevant clip indices $R^{(j)} \subseteq \{1,\dots,T\}$ by matching $q^{(j)}$ against the stored keys $\{K_i\}$ using a sentence embedding model $E(\cdot)$ and similarity scoring.
For each retrieved index $i \in R^{(j)}$, the agent collects the corresponding episodic record as an evidence item $e_i = (v_i, d_i)$,
where $d_i$ is used during the iterative retrieval, and $v_i$ is reserved for generating the final answer $A$. 
It updates the accumulated evidence set as
\begin{equation}
\label{eq:evidence_set}
\mathcal{E}^{(j)} = \mathcal{E}^{(j-1)} \cup \{e_i \mid i \in R^{(j)}\}.
\end{equation}
After each retrieval round, the agent assesses whether the currently collected evidence is sufficient to answer $Q$. If not, it refines the retrieval query and repeats retrieval. Once sufficient evidence is obtained, the agent leverages the solver’s reasoning to integrate evidence across retrieved episodes and infer the relationships required to produce $A$.

\subsection{Multi-Key Memory and Retrieval}

\label{sec:multi-key}
\ours{} defers query-specific relation reasoning to the solver at inference time, so retrieval becomes the key bottleneck, as the solver can only reason over relationships present in the retrieved evidence $\mathcal{E}$.
In long video QA, a clip can be relevant via different cues, such as actions, spoken mentions, object interactions and state changes, or coarse scene context; collapsing these cues into a single textual index is often brittle.
Therefore, instead of the standard single-key indexing, we represent each record $(v_i, d_i)$ with a \emph{set} of complementary keys $K_i$ and use late-interaction matching~\cite{khattab2020colbert, reddy2025video, wan2025clamr} to favor high-recall retrieval.

\subsubsection{Multi-Key Memory.}
The representation is explicitly formulated as a combination of four distinct categories, defined as
\begin{equation}
K_i = \{k^{evt}_i, k^{dial}_i, k^{obj}_i, k^{sum}_i\}.
\end{equation}
\begin{itemize}
  \item Event~/~Action Key ($k^{evt}_i$): Captures observable physical actions and interactions between entities.
  \item Dialogue~/~Mention Key ($k^{dial}_i$): Records spoken content, exact words, or commands mentioned in the audio track, providing crucial linguistic context.
  \item Object Key ($k^{obj}_i$): Describes specific items being handled, requested, or moved, along with their state transitions.
  \item Summary Key ($k^{sum}_i$): A compact keyword-style abstraction capturing the core narrative and the coarse information of the overall clip.
\end{itemize}
This decoupled representation ensures that each clip can be matched from diverse perspectives, effectively handling varying query intents.

\subsubsection{Maximum Similarity Retrieval.}
At retrieval round $j$, given the retrieval query $q^{(j)}$, we embed $q^{(j)}$ and each key $k \in K_i$ using $E(\cdot)$ and define the clip relevance by the maximum cosine similarity:
\begin{equation}
\label{eq:clip_score}
S_i^{(j)} = \max_{k \in K_i} \text{sim}\!\big(E(q^{(j)}),\, E(k)\big),
\end{equation}
where $\text{sim}(\cdot, \cdot )$ denotes the cosine similarity score between two embeddings.
We then select the top $N$ clips:
\begin{equation}
\label{eq:topn_retrieval}
R^{(j)} = \mathrm{TopN}\big(\{S_i^{(j)}\}_{i=1}^{T}\big).
\end{equation}
This maximum similarity formulation improves recall by allowing each clip to match the query through its best-aligned aspect.

\subsection{Temporal Expansion via Neighbor Filtering}

\label{sec:NF}
Videos exhibit strong temporal locality: the evidence required to answer a query often spans multiple nearby moments rather than a single 30-second clip. While short clips enable lightweight indexing, retrieving only the anchor clip can miss necessary preconditions or follow-up context, limiting the solver's ability to infer query-specific relationships.

\subsubsection{Neighbor Filtering.}
We exploit this inductive bias with a minimal temporal structure.
Instead of pre-computing multi-granularity hierarchies, we attach a local neighborhood of temporally adjacent clips to each retrieved anchor on demand, effectively forming the simplest graph over the timeline. 
This query-driven expansion enriches the retrieved evidence at low cost, allowing a powerful solver to compose higher-level temporal and semantic relations in a query-aware manner at inference time.

For each retrieved index $i\in R^{(j)}$ in round $j$, we define a symmetric temporal window
\begin{equation}
\label{eq:nbhd}
W_i = \{t \in \{1,...,T\}|i-\Delta \leq t \leq i+\Delta\},
\end{equation}
where $\Delta$ is an integer radius indicating 
the number of adjacent clips on each side 
(e.g., $\Delta{=}2$ spans $\pm$1 minute under a 30-second clip setting).

The full textual context for neighbor filtering at round $j$ is obtained by concatenating the dense captions within the neighborhood of each retrieved index:
\begin{equation}
D^{(j)} = \operatorname{Concat}\big(\{d_t|t\in W_i, i \in R^{(j)}\}\big).
\end{equation}
We then use the solver $\mathcal{G}$ as a query-aware filter to distill only the information relevant to $Q$ from this expanded context: $I^{(j)} = \mathcal{G}_{\mathrm{filter}}(Q, D^{(j)})$,
where $\mathcal{G}_{\mathrm{filter}}$ denotes the solver equipped with a filtering prompt.
The resulting $I^{(j)}$ is appended to the evidence set $\mathcal{E}^{(j)}$ 
and used for subsequent query-aware reasoning.
\section{Experiments}

\subsection{Benchmarks}

We evaluate \ours{} against existing memory-based models on three long-video benchmarks, formulating all tasks as multiple-choice questions (MCQs) and using QA accuracy as the primary evaluation metric.

\noindent\textbf{EgoLifeQA~\cite{egolife}}: An ultra-long egocentric dataset averaging 44.3 hours per video, capturing six individuals over seven days with 500 QA pairs. It features five QA types: EntityLog and EventRecall test precise factual grounding of objects and events, while HabitInsight, RelationMap, and TaskMaster evaluate semantic reasoning over human behaviors, interactions, and future planning. 
\noindent\textbf{LVBench~\cite{lvbench}}: A benchmark for hour-long video understanding with an average duration of 1.12 hours and a maximum of 2.33 hours. It includes 1,549 questions across six skill categories: Entity Recognition (ER) and Event Understanding (EU) for tracking and classification; Temporal Grounding (TG) and Key Information Retrieval (KIR) for locating specific moments and details; and Summarization (Sum) and Reasoning (Rea) for synthesizing global content and causal inference. Among these, ER and EU account for the largest proportion.
\noindent\textbf{Video-MME~\cite{videomme}}: This is a comprehensive benchmark for evaluating the general video understanding capabilities of MLLMs. It is categorized into Short, Medium, and Long subsets based on video duration. In our experiments, we use only the Long subset (30–60 mins), which comprises 900 questions from 300 videos, with three questions per video. The benchmark covers 12 question types, with Object Reasoning, Action Reasoning, and Information Synopsis being the most prevalent in the Long subset.

\begin{table}[!t]
\centering
\caption{
Comparison of \ours{} against various baselines on EgoLifeQA~\cite{egolife}. Baseline results are from~\cite{worldmm,egagent}. \textbf{Bold} and \underline{underline} denote the best and second-best scores, respectively. Solvers are indicated per model (pre-trained models report weights only).}
\small
\renewcommand{\arraystretch}{1.15}
\label{tab:table_1_EgoLifeQA}
\resizebox{0.9\linewidth}{!}{%
\begin{tabular}{lcccccc} 
\toprule
\textbf{Model} & \textbf{EntityLog} & \textbf{EventRecall} & \textbf{HabitInsight} & \textbf{RelationMap} & \textbf{TaskMaster} & \textbf{Avg} \\ \midrule \midrule
\multicolumn{7}{l}{\textit{\textbf{MLLMs (Uniform Sampling)}}} \\ 
Qwen3-VL-8B~\cite{qwen3-vl} & 35.2 & 30.2 & 39.3 & 46.4 & 46.0 & 38.6 \\
Gemini 2.5 Pro~\cite{gemini2.5} & 43.2 & 40.5 & 41.0 & 55.2 & 52.4 & 46.4 \\
GPT-5~\cite{gpt5} & 47.2 & 42.1 & 47.5 & 53.6 & 55.6 & 48.6 \\ \midrule
\multicolumn{7}{l}{\textit{\textbf{Hierarchical Memory Based}}} \\
EgoRAG~\cite{egolife} (GPT-5) & 40.0 & 56.3 & 62.3 & 54.4 & 52.4 & 52.0 \\  
Ego-R1~\cite{egor1} (3B) & 51.2 & 53.2 & 63.9 & 50.4 & 50.8 & 53.0 \\ \midrule
\multicolumn{7}{l}{\textit{\textbf{Graph Memory Based}}} \\ 
LightRAG~\cite{lightrag} (GPT-5) & 40.8 & 48.4 & 67.2 & 50.4 & 44.4 & 48.8 \\
HippoRAG~\cite{hipporag} (GPT-5) & 48.8 & 60.3 & 70.5 & 60.8 & 66.7 & 59.6 \\
Video-RAG~\cite{video-rag} (GPT-5) & 49.6 & 56.3 & 67.2 & 55.2 & 54.0 & 55.4 \\ 
HippoMM~\cite{hippomm} (GPT-5) & 45.6 & 53.2 & 70.5 & 55.2 & 58.7 & 54.6 \\
M3-Agent~\cite{m3-agent} (7B) & 44.4 & 54.8 & 62.3 & 56.8 & 54.0 & 53.5 \\
EGAgent~\cite{egagent} (Gemini 2.5 Pro) & 54.4 & 57.1 & 60.3 & 62.4 & \textbf{74.6} & 57.5 \\  
WorldMM~\cite{worldmm} (Qwen3-VL-8B) & 49.6 & 56.4 & 63.9 & 58.4 & 58.7 & 56.4 \\
WorldMM~\cite{worldmm} (GPT-5) & \underline{62.4} & \underline{64.3} & \textbf{75.4} & 62.4 & 71.4 & \underline{65.6} \\ \midrule
\multicolumn{7}{l}{\textit{\textbf{Ours}}} \\ 
\rowcolor[gray]{0.95} 
\ours{} (Qwen3-VL-8B) & 43.2 & 54.0 & 67.2 & 53.6 & 65.1 & 54.2 \\
\rowcolor[gray]{0.95} 
\ours{} (Gemini 2.5 Pro) & 60.8 & 61.1 & 65.6 & \underline{65.6} & \underline{73.0} & 64.2 \\
\rowcolor[gray]{0.95} 
\ours{} (GPT-5) & \textbf{67.2} & \textbf{70.6} & \underline{73.8} & \textbf{74.4} & 71.4 & \textbf{71.2} \\ \bottomrule
\end{tabular}%
}
\end{table}

\subsection{Implementation Details}

To evaluate our framework, we applied \ours{} in Qwen3-VL-8B~\cite{qwen3-vl}, GPT-5~\cite{gpt5}, and Gemini 2.5 Pro~\cite{gemini2.5}. Following the previous work~\cite{worldmm}, we employ GPT-5-mini~\cite{gpt5} to generate dense captions for each 30-second clip. We also use the same model to generate episodic multi-key for each segment.
For EgoLifeQA~\cite{egolife} and Video-MME(Long)~\cite{videomme}, we incorporate ASR transcripts as an additional modality during captioning~\cite{egagent, worldmm}. For LVBench~\cite{lvbench}, which does not rely on dialogue or speech, we use only visual frames for captioning and exclude the dialogue key. The prompts used for multi-key extraction and neighbor filtering, along with further implementation details, are provided in the Appendix.

\subsection{Main Results}

\begin{table}[!t]
\centering
\caption{Comparison of MERIT against various baselines on LVBench~\cite{lvbench} and Video-MME(L)~\cite{videomme} benchmarks. Baseline results are from~\cite{worldmm, egagent}. Solvers are indicated per model (pre-trained models report weights only).}
\small
\renewcommand{\arraystretch}{1.2} 
\label{tab:table_2_LV_MME}
\resizebox{0.6\linewidth}{!}{
\begin{tabular}{lcc}
\toprule
\textbf{Model} & \textbf{LVBench} & \textbf{Video-MME(L)} \\ \midrule \midrule

\multicolumn{3}{l}{\textit{\textbf{MLLMs (Uniform Sampling)}}} \\ 
Gemini 2.5 Pro~\cite{gemini2.5} & 57.0 & 55.7 \\
GPT-5~\cite{gpt5} & 60.4 & 74.3 \\ \midrule

\multicolumn{3}{l}{\textit{\textbf{Hierarchical Memory Based}}} \\ 
EgoRAG~\cite{egolife} (GPT-5) & 32.2 & 41.1 \\
Ego-R1~\cite{egor1} (3B) & 34.1 & 42.7 \\ \midrule

\multicolumn{3}{l}{\textit{\textbf{Graph Memory Based}}} \\ 
LightRAG~\cite{lightrag} (GPT-5) & 30.4 & 46.6 \\
HippoRAG~\cite{hipporag} (GPT-5) & 54.0 & 52.1 \\
Video-RAG~\cite{video-rag} (GPT-5) & 33.1 & 55.4 \\ 
HippoMM~\cite{hippomm} (GPT-5) & 38.2 & 41.6 \\
M3-Agent~\cite{m3-agent} (7B)& 49.3 & 55.3 \\
EGAgent~\cite{egagent} (Gemini 2.5 Pro) & - & 74.1 \\
WorldMM~\cite{worldmm} (GPT-5) & 61.9 & 76.6 \\ \midrule

\multicolumn{3}{l}{\textit{\textbf{Ours}}} \\ 
\rowcolor[gray]{0.95}
\ours{} (Gemini 2.5 Pro) & - & 76.3 \\
\rowcolor[gray]{0.95}
\ours{} (GPT-5) & \textbf{71.8} & \textbf{77.7} \\ \bottomrule
\end{tabular}%
}
\end{table}

\noindent\textbf{Results on Ultra-Long Video QA.}
Table~\ref{tab:table_1_EgoLifeQA} presents the evaluation on the ultra-long benchmark EgoLifeQA. Uniform-sampling MLLMs exhibit the lowest overall performance due to strict context length limits. This confirms that processing full ultra-long videos is computationally prohibitive and inaccurate, thereby necessitating memory-based architectures. \ours{} outperforms existing memory-based approaches and establishes a new state-of-the-art. Using GPT-5, our framework achieves an average accuracy of 71.2\%, yielding a +5.6\% improvement over the previous SOTA method. 

An analysis of individual QA types highlights the strengths of \ours{} in fine-grained categories such as EntityLog and EventRecall. This proves that our simple key representation prevents information loss during retrieval, passing intact raw evidence directly to the solver. Furthermore, the significant +12.0\% improvement in RelationMap proves that expanding the temporal radius during inference successfully captures query-relevant semantic information without the heavy overhead of pre-structuring semantic memory. This superiority is consistent across advanced models; utilizing Gemini 2.5 Pro yields a +6.7\% gain over the best graph-memory baseline. Consequently, these results indicate that \ours{} effectively leverages advanced reasoning to maximize performance.

\noindent\textbf{Results on Hour-Long Video QA.}
Table~\ref{tab:table_2_LV_MME} demonstrates that \ours{} also establishes new SOTA results on hour-long benchmarks. Specifically, on LVBench, our GPT-5 setting achieves 71.8\%, outperforming the prior SOTA by a significant margin of +9.9\%. Unlike the day-long setting, uniform-sampling MLLMs show competitive performance on these shorter videos (averaging 0.69 hours for Video-MME(L) and 1.12 hours for LVBench), occasionally surpassing existing memory-based models. However, as video duration increases to over an hour in LVBench, the performance gap between uniform sampling and \ours{} widens significantly. For instance, with GPT-5, the gap expands from +3.4\% on Video-MME(L) to an impressive +11.4\% on LVBench. This confirms that while MLLMs can manage moderately long contexts, their scalability strictly limits them in longer settings, highlighting the robustness and the efficacy of our memory framework for arbitrarily long videos.

\subsection{Ablation Studies}
We conduct ablation studies to evaluate the individual components of \ours{}. All ablation experiments are conducted on the EgoLifeQA~\cite{egolife} benchmark. For the backbone QA solvers, we use both open-source and proprietary models. To analyze retrieval performance, we additionally measure \textit{Hit Rate}. Given a query and its corresponding target timestamp, a hit is scored as 1 if the target time falls within the retrieved memory interval, and 0 otherwise. The Hit Rate is the average of hit scores across all queries.

\begin{table}[!t]
  \centering
  \caption{Ablation study on Neighbor Filtering (NF) across two solver models. \textit{Target Time Oracle} provides ground-truth temporal segments as an upper bound, while \ours{} uses retrieved segments.}
  \normalsize
  \renewcommand{\arraystretch}{1.3} 
  \label{tab:table_3_NF}
  \resizebox{1\textwidth}{!}{%
    \begin{tabular}{l|c|c|c|c|c|c|c|c}
      \toprule
      \textbf{EXP} & \textbf{NF} & \textbf{EntityLog} & \textbf{EventRecall} & \textbf{HabitInsight} & \textbf{RelationMap} & \textbf{TaskMaster} & \textbf{Avg} & \textbf{Hit Rate} \\ \midrule
      \multirow{2}{*}{Target Time Oracle (Qwen3-VL-8B~\cite{qwen3-vl})} & \ding{55} & \textbf{60.0} & \textbf{69.8} & \textbf{70.5} & 54.4 & 73.0 & 64.0 & 1.0 \\
      & \cellcolor[gray]{0.9}\checkmark & \cellcolor[gray]{0.9}56.8 & \cellcolor[gray]{0.9}66.7 & \cellcolor[gray]{0.9}68.9 & \cellcolor[gray]{0.9}\textbf{67.2} & \cellcolor[gray]{0.9}\textbf{76.2} & \cellcolor[gray]{0.9}\textbf{65.8} & \cellcolor[gray]{0.9}1.0 \\ \cline{1-9}
      \multirow{2}{*}{\ours{} (Qwen3-VL-8B~\cite{qwen3-vl})} & \ding{55} & 41.6 & \textbf{54.0} & 50.8 & 42.4 & 57.1 & 48.0 & 0.35 \\
      & \cellcolor[gray]{0.9}\checkmark & \cellcolor[gray]{0.9}\textbf{43.2} & \cellcolor[gray]{0.9}\textbf{54.0} & \cellcolor[gray]{0.9}\textbf{67.2} & \cellcolor[gray]{0.9}\textbf{53.6} & \cellcolor[gray]{0.9}\textbf{65.1} & \cellcolor[gray]{0.9}\textbf{54.2} & \cellcolor[gray]{0.9}\textbf{0.42} \\ \midrule
      \multirow{2}{*}{Target Time Oracle (GPT-5~\cite{gpt5})} & \ding{55} & \textbf{82.4} & \textbf{89.7} & \textbf{85.3} & 73.6 & 82.5 & 82.4 & 1.0 \\
      & \cellcolor[gray]{0.9}\checkmark & \cellcolor[gray]{0.9}80.8 & \cellcolor[gray]{0.9}88.1 & \cellcolor[gray]{0.9}\textbf{85.3} & \cellcolor[gray]{0.9}\textbf{76.8} & \cellcolor[gray]{0.9}\textbf{88.9} & \cellcolor[gray]{0.9}\textbf{83.2} & \cellcolor[gray]{0.9}1.0 \\ \cline{1-9}
      \multirow{2}{*}{\ours{} (GPT-5~\cite{gpt5})}  & \ding{55} & \textbf{68.0} & 69.8 & 68.9 & 73.6 & 68.3 & 70.0 & 0.40 \\
      & \cellcolor[gray]{0.9}\checkmark & \cellcolor[gray]{0.9}67.2 & \cellcolor[gray]{0.9}\textbf{70.6} & \cellcolor[gray]{0.9}\textbf{73.8} & \cellcolor[gray]{0.9}\textbf{74.4} & \cellcolor[gray]{0.9}\textbf{71.4} & \cellcolor[gray]{0.9}\textbf{71.2} & \cellcolor[gray]{0.9}\textbf{0.56} \\ \bottomrule
    \end{tabular}%
}
\end{table}

\subsubsection{Effect of Neighbor Filtering.}
Table~\ref{tab:table_3_NF} evaluates the impact of expanded temporal context on QA performance. Neighbor Filtering (NF) is a mechanism that dynamically expands the temporal radius around a retrieved clip at inference time to incorporate surrounding information. In the absence of NF, the solver is strictly limited to the memory of the single retrieved 30-second clip, without any adjacent context.

To establish an upper bound for this single-clip setting, we evaluate a ``Target Time Oracle'' configuration, which directly provides the solver with frames and captions from the ground-truth 30-second clip. This oracle setting reveals a substantial performance gap compared to prior state-of-the-art models, confirming that precise retrieval remains the primary bottleneck in ultra-long video QA.

Applying NF mitigates this bottleneck by expanding the temporal context during inference, which significantly improves both the hit rate and overall accuracy. Specifically, the expanded temporal scope drives consistent performance gains in QA categories that inherently require a broader semantic context, such as HabitInsight, RelationMap, and TaskMaster.

Notably, these improvements are also consistently observed in the ``Target Time Oracle w/ NF'' setting. This indicates that even with perfect temporal localization, a single isolated 30-second segment often lacks sufficient episodic context to resolve complex queries. By incorporating information from surrounding clips, NF provides the necessary temporal and relational cues to capture a complete semantic understanding. Consequently, this on-demand expansion yields significant performance gains without incurring the massive computational overhead of additional pre-processing.

\begin{table}[t]
  \centering
  \small
  \renewcommand{\arraystretch}{1.3}
  \begin{tabular}{@{}c@{\hspace{0.03\columnwidth}}c@{}}
    \begin{minipage}[t]{0.485\columnwidth}
      \vspace{0pt}\centering
      \caption{Performance analysis based on key combinations with Qwen3-VL-8B~\cite{qwen3-vl}. Darker colors indicate higher accuracy.}
      \label{tab:table_4_key}
      \resizebox{0.95\linewidth}{!}{%
        \definecolor{best}{HTML}{D9822B}   
\definecolor{high}{HTML}{EBAE75}    
\definecolor{medium}{HTML}{F2C294}  
\definecolor{low}{HTML}{F9E1C5}    

\newcommand{\mybox}[2]{\cellcolor{#1}#2}

\begin{tabular}{c|c|c|c|c|c|c}
\toprule
 \textbf{$N$ Key} & \textbf{Event} & \textbf{Dial} & \textbf{Object} & \textbf{Sum} & \textbf{Acc} & \textbf{Avg} \\ \midrule
\multirow{4}{*}{1 key} & \checkmark & & & & 44.6 & \multirow{4}{*}{46.4} \\ \cline{2-6} 
 & & \checkmark & & & 44.8 & \\ \cline{2-6} 
 & & & \checkmark & & 46.8 & \\ \cline{2-6} 
 & & & & \checkmark & 49.4 & \\ \midrule
\multirow{6}{*}{2 key} & \checkmark & \checkmark & & & 49.6 & \multirow{6}{*}{49.4} \\ \cline{2-6} 
 & \checkmark & & \checkmark & & 47.0 & \\ \cline{2-6} 
 & \checkmark & & & \checkmark & 49.0 & \\ \cline{2-6} 
 & & \checkmark & \checkmark & & 49.2 & \\ \cline{2-6} 
 & & \checkmark & & \checkmark & \mybox{low}{51.2} & \\ \cline{2-6} 
 & & & \checkmark & \checkmark & 50.6 & \\ \midrule
\multirow{4}{*}{3 key} & \checkmark & \checkmark & \checkmark & & 49.8 & \multirow{4}{*}{51.1} \\ \cline{2-6} 
 & \checkmark & \checkmark & & \checkmark & \mybox{high}{52.0} & \\ \cline{2-6} 
 & \checkmark & & \checkmark & \checkmark & 50.8 & \\ \cline{2-6} 
 & & \checkmark & \checkmark & \checkmark & \mybox{medium}{51.6} & \\ \midrule
4 key & \checkmark & \checkmark & \checkmark & \checkmark & \mybox{best}{\textbf{54.2}} & 54.2 \\ \bottomrule
\end{tabular}

      }
    \end{minipage}
    &
    \begin{minipage}[t]{0.485\columnwidth}
      \vspace{0pt}\centering
      \caption{Retrieval accuracy comparison with baseline models using GPT-5~\cite{gpt5}. The ($^*$) denotes values obtained from our reproduction of the baseline models under the same evaluation settings.}
      \label{tab:table_5_hit}
      \resizebox{0.7\linewidth}{!}{%
        \begin{tabular}{c|c|c}
\toprule
\textbf{EXP} & \textbf{QA Acc} & \textbf{Hit Rate} \\ \midrule
EgoRAG$^*$ & 49.7 & 0.15 \\ \hline
WorldMM$^*$ & 63.2 & 0.53 \\ \hline
\ours{} & \textbf{71.2} & \textbf{0.56} \\ \bottomrule

\end{tabular}
      }
      \par\vspace{0.5em}
      \setlength{\tabcolsep}{4pt}
  \renewcommand{\arraystretch}{1.4}
  \scriptsize
  \captionsetup{skip=2pt}
\caption{Comparison of memory construction efficiency with reproduced baseline models($^*$) using GPT-5~\cite{gpt5}.}
\label{tab:table_6_token}
\resizebox{\linewidth}{!}{%
\begin{tabular}{c|ccc}
\toprule
\multicolumn{1}{c}{\raisebox{0.5\height}{\textbf{EXP}}} & \multicolumn{1}{|c}{\textbf{\shortstack{LLM Call\\Count}}} & \multicolumn{1}{c}{\textbf{\shortstack{Input\\Tokens}}} & \multicolumn{1}{c}{\textbf{\shortstack{Output\\Tokens}}} \\
\midrule
EgoRAG$^*$ & \textbf{339}    & 1,049K & \textbf{247K}  \\ \hline
WorldMM$^*$ & 23,731 & 6,860K & 2,472K \\ \hline
\ours{} & 6,223  & \textbf{815K}   & 288K  \\
\bottomrule
\end{tabular}
}
 
    \end{minipage}
  \end{tabular}
\end{table}

\subsubsection{Effect of Multi-Key Combinations.}
Table~\ref{tab:table_4_key} details the performance variations across different configurations of our multi-key representation using the Qwen3-VL-8B\cite{qwen3-vl}. Specifically, our framework extracts four distinct keys per clip: event, dialogue, object, and summary. This ablation study evaluates the impact of these individual keys, their various combinations, and the effect of scaling the total number of keys utilized for retrieval.

Under the single-key setting, the summary key outperforms the other individual keys by a margin of up to +4.8\%. Instead of focusing on specific aspects, the summary key captures coarse, abstract context, serving as a robust primary anchor for retrieval.

Furthermore, combining multiple keys leads to a steady increase in overall accuracy. While intermediate multi-key combinations show minor variances, the distinct keys function in a highly complementary manner. Consequently, utilizing all four keys simultaneously yields the peak accuracy of 54.2\%. These findings empirically validate our premise: maintaining a decoupled multi-key representation ensures that each video clip can be accurately matched from diverse semantic perspectives, thereby effectively handling varying query intents.

\subsubsection{Retrieval Accuracy vs. Baselines.}
To investigate the correlation between retrieval accuracy and overall QA performance, we evaluate the hit rate of reproduced models representing distinct memory architectures (Table~\ref{tab:table_5_hit}). These baselines are built on the same 30-second dense captions as \ours{}, produced by the same captioner~\cite{gpt5}, using their official code for controlled comparison.

Compared to the hierarchical memory of EgoRAG~\cite{egolife} and the graph memory of WorldMM~\cite{worldmm}, \ours{} employs the simplest memory key structure yet achieves both the highest hit rate and the highest overall accuracy. This direct alignment confirms that successful information retrieval is critical for QA performance. Unlike \ours{} and EgoRAG, which retrieve in a fixed 30-second granularity, WorldMM retrieves clips across multi-scale temporal windows, including up to 1 hour segments. While retrieving such extensive temporal windows inherently increases the probability of capturing the target timestamp, our framework still achieves a higher hit rate using only a single granularity. In our reproduction, evaluating WorldMM solely with 30-second clips yields a hit rate of 0.31. These results demonstrate that precise, fine-grained retrieval capabilities are closely correlated with maximizing overall QA performance.

\subsubsection{Memory Construction Efficiency.}
In Table~\ref{tab:table_6_token}, we evaluate the memory construction cost. Since all methods use the same 30-second captions, we compare memory construction efficiency by measuring the total number of caption tokens consumed by each method, excluding the common initial captioning step.

Compared to WorldMM~\cite{worldmm}, \ours{} demonstrates a significant efficiency advantage, achieving an \textbf{8.4$\times$} reduction in input tokens and an \textbf{8.6$\times$} reduction in output tokens. 
WorldMM's pipeline generates captions across four granularities, applies OpenIE to each, and continuously updates multi-level semantic graphs, incurring 23.7K LLM calls and millions of input/output tokens. In contrast, \ours{} builds its memory in a single pass: each clip is captioned once and indexed with lightweight multi-keys, which eliminates the repeated generation and graph-update steps.

Furthermore, while EgoRAG~\cite{egolife} requires fewer LLM calls by batching inputs within its hierarchical structure, \ours{} consumes 22\% fewer input tokens (815K vs. 1.05M). This efficiency is achieved because \ours{} processes each caption exactly once, avoiding the redundant re-reading of intermediate summaries inherent to EgoRAG's hierarchy levels. Although \ours{} generates slightly more output tokens than EgoRAG, this is a deliberate design choice: rather than heavily compressing information into single summaries, \ours{} generates fine-grained multi-keys per clip, which is critical for preserving detailed temporal context and enabling highly accurate retrieval.

\definecolor{myblue}{HTML}{D9E2F3}
\definecolor{mypink}{HTML}{F2DCDB}

\begin{table}[t]
  \centering
  \small
  \renewcommand{\arraystretch}{1.3}
  \caption{Impact of the retrieved clip count (Top $N$) across both open-source and proprietary QA solvers. Colors denote higher performance within each individual solver.}
  \label{tab:table_7_topN}
  \makebox[\linewidth][c]{%
    \resizebox{0.9\linewidth}{!}{%
      \begin{tabular}{c|c|c|c|c|c|c|c}
        \toprule
        \textbf{Solver} & \textbf{Top $N$} & \textbf{EntityLog} & \textbf{EventRecall} & \textbf{HabitInsight} & \textbf{RelationMap} & \textbf{TaskMaster} & \textbf{Avg} \\ \midrule
        \multirow{2}{*}{Qwen3-VL-4B~\cite{qwen3-vl}} & \cellcolor{myblue}5 & 41.6 & 47.6 & 63.9 & 37.6 & 55.6 & \cellcolor{myblue}\textbf{46.6} \\ \cline{2-8}
        & 10 & 37.6 & 53.2 & 59.0 & 39.2 & 52.4 & 46.4 \\ \midrule
        \multirow{2}{*}{Qwen3-VL-8B~\cite{qwen3-vl}} & \cellcolor{myblue}5 & 43.2 & 54.0 & 67.2 & 53.6 & 65.1 & \cellcolor{myblue}\textbf{54.2} \\ \cline{2-8}
        & 10 & 44.0 & 57.9 & 63.9 & 40.8 & 68.3 & 52.2 \\ \midrule \midrule
        \multirow{2}{*}{Gemini 2.5 pro~\cite{gemini2.5}} & 5 & 53.6 & 62.7 & 60.7 & 65.6 & 66.7 & 61.4 \\ \cline{2-8}
        & \cellcolor{mypink}10 & 60.8 & 61.1 & 65.6 & 65.6 & 73.0 & \cellcolor{mypink}\textbf{64.2} \\ \midrule
        \multirow{2}{*}{GPT-5~\cite{gpt5}} & 5 & 62.4 & 67.5 & 73.8 & 68.0 & 69.8 & 67.4 \\ \cline{2-8}
        & \cellcolor{mypink}10 & 67.2 & 70.6 & 73.8 & 74.4 & 71.4 & \cellcolor{mypink}\textbf{71.2} \\ \bottomrule
      \end{tabular}%
    }
  }
\end{table}

\subsubsection{Effect of Retrieved Clip Quantity Across Solver Capacities.}
Table~\ref{tab:table_7_topN} investigates the effect of the number of retrieved clips (Top $N$) across QA solvers with varying capacities. We observe a distinct divergence in performance trends dependent on the solver's inherent capabilities. For open-source models~\cite{qwen3-vl}, restricting retrieval to Top 5 clips yields optimal results, whereas expanding to Top 10 degrades performance. This suggests that models with limited reasoning capacity struggle to effectively consolidate information from extended contexts, often becoming confused by the increased noise during inference. 

Conversely, proprietary models~\cite{gemini2.5,gpt5} demonstrate substantial improvements when provided with a larger pool of clips. Increasing to Top 10 boosts the accuracy of Gemini 2.5 Pro (61.4\% to 64.2\%) and GPT-5 (67.4\% to 71.2\%). This suggests that high-capacity solvers possess the noise-tolerance required to effectively filter query-aware evidence from expanded context. Rather than being hindered by redundant information in additional temporal windows, they distill critical semantic cues from extended contexts to enhance reasoning accuracy.

\section{Qualitative Results}

\begin{figure}[t]
  \centering
  \setlength{\abovecaptionskip}{4pt}
  \setlength{\belowcaptionskip}{2pt}
  \includegraphics[width=1\linewidth]{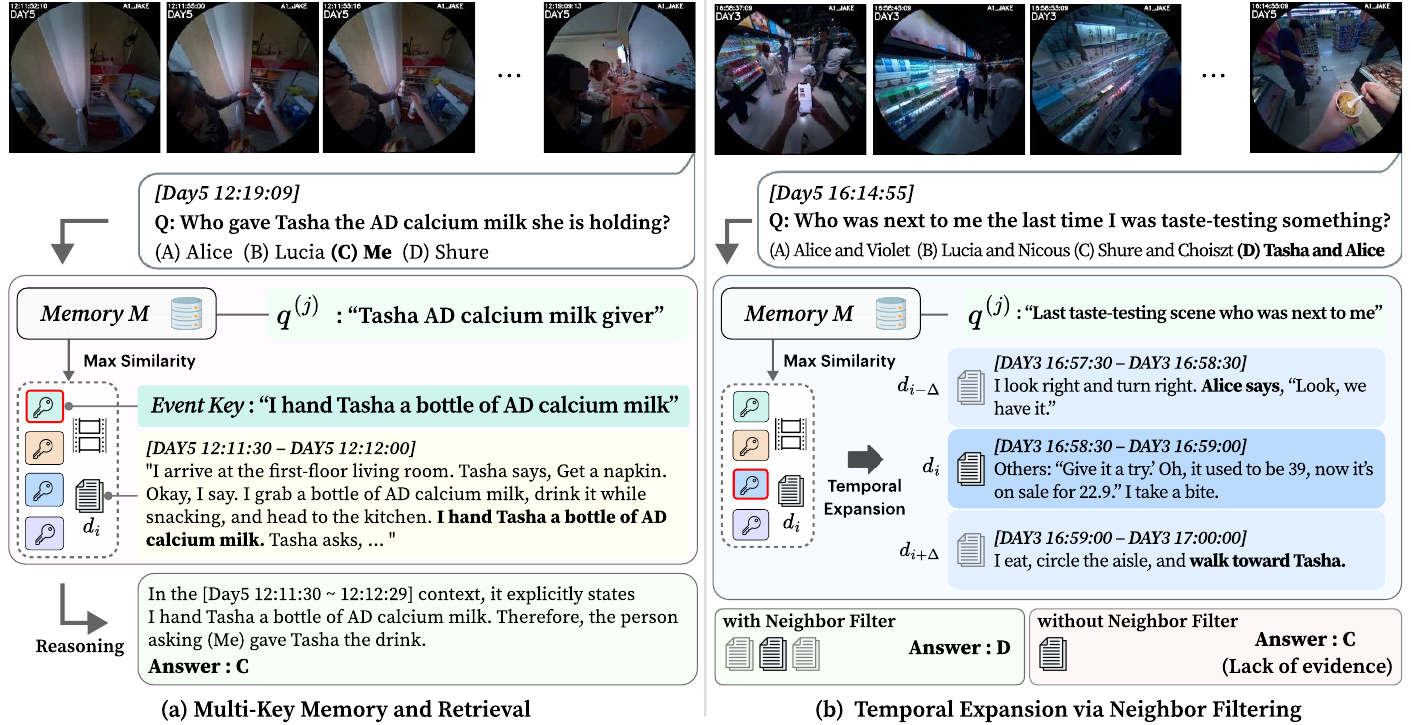}
  \caption{Qualitative results of \ours{} on EgoLifeQA~\cite{egolife}. (a) The event key precisely matches the query, enabling retrieval of the exact target segment. (b) While both settings retrieve the correct segment, neighbor filter supplies the surrounding $\pm\Delta$ context necessary to resolve relational queries that cannot be answered from a single clip alone.}
  \label{fig:fig3}
\end{figure}

\subsubsection{Effect of Multi-Key Memory.}
Fig.~\ref{fig:fig3} (a) illustrates an example from the EntityLog category of EgoLifeQA~\cite{egolife}. The query requires identifying the specific moment an object was handed over. Our multi-key memory matches this query to the stored event key: \emph{``I hand Tasha a bottle of AD calcium milk.''} By retrieving the exact target segment associated with this key, our retrieval mechanism provides the solver with the visual evidence necessary to identify the target entity. This successful retrieval of the fine-grained event leads to the correct answer.

\subsubsection{Effect of Neighbor Filtering.}
Fig.~\ref{fig:fig3} (b) presents a RelationMap case from EgoLifeQA~\cite{egolife}, requiring an understanding of interpersonal relationships and interactions. The query asks about an event from two days before the query time, where multiple individuals are present. While multi-key matching accurately retrieves the target segment, a single clip lacks the context to identify specific individuals.
Neighbor filtering resolves this by expanding the temporal context, allowing the query-aware filter to extract the details: \emph{``I was taste-testing and next to me were Tasha and Alice.''} This confirms that while multi-key indexing localizes anchor clips, neighbor filtering provides the broader situational context required for relational queries—without relying on pre-built semantic memories.
\section{Conclusion}
We present \ours{}, a simple yet effective framework for ultra-long video understanding that preserves fine-grained episodic evidence through multi-key indexing and enriches retrieved context via query-time neighbor filtering. Despite its simplicity, \ours{} achieves state-of-the-art performance across multiple benchmarks, suggesting that deferring semantic composition to inference time is more effective than query-agnostic preprocessing. We hope \ours{} serves as a practical and scalable foundation for future work on ultra-long video understanding.

\section*{Acknowledgements}
This work was supported by Institute of Information \& Communications Technology Planning \& Evaluation (IITP) grant funded by the Korea government (MSIT) (No. RS-2020-II201361, Artificial Intelligence Graduate School Program (Yonsei University)), National Research Foundation of Korea (NRF)~(RS-2025-00554790), and No.RS-2022-II220124, Development of Artificial Intelligence~Technology for Self-Improving Competency-Aware Learning Capabilities.

\bibliographystyle{splncs04}
\bibliography{main}

\clearpage
\appendix
\renewcommand{\theHsection}{appendix.\Alph{section}}
\renewcommand{\theHsubsection}{appendix.\Alph{section}.\arabic{subsection}}
\section*{Supplementary Materials} 
This supplementary material provides additional details and analyses that complement the main paper. We first describe additional implementation details of \ours{} (Section~\ref{appendix:impl}), followed by detailed performance comparisons on LVBench and Video-MME(Long) (Section~\ref{appendix:comparison}). We then present additional ablations and analyses, including the retrieval round distribution, a comparison with a flat retrieval baseline, and the effect of the neighbor filtering window size (Section~\ref{appendix:ablation}). Next, we discuss the solver dependency, intelligence allocation, and future directions of our framework (Section~\ref{appendix:discussion}). We further provide additional qualitative analyses, including comparisons with memory-based baselines and failure cases (Section~\ref{appendix:qualitative}). Lastly, we list all prompts used throughout our pipeline (Section~\ref{appendix:prompts}).

\section{Additional Implementation Details}\label{appendix:impl}

\noindent\textbf{Data Preprocessing and Retrieval Pipeline.}
To generate the textual captions, we partition the input video into 30-second clips and uniformly sample frames at 1 FPS. During memory construction, the model is prompted to extract four distinct types of keys simultaneously, relying solely on these generated captions without visual inputs. For the retrieval process, we compute embeddings for both the natural language query and the stored keys using the Qwen3-Embedding-4B~\cite{qwen3embedding}. Once the top-$N$ anchor clips are retrieved, we apply temporal expansion via neighbor filtering. The concatenated textual context from these local neighborhoods is processed in a single batch by the solver with a filtering prompt to distill query-relevant information. Consequently, each retrieval round requires exactly two Large Language Models(LLM) calls: one for assessing evidence sufficiency and formulating the query, and another for neighbor filtering. Once the retrieval loop terminates, one final LLM call is executed during the ultimate QA stage. In our experimental setup, all inference-time roles, including the retrieval agent and the solver, are served by a single identical MLLM backbone $\mathcal{G}$. All specific LLM prompts used in this pipeline are provided in Section~\ref{sec:prompts}.

\noindent\textbf{Frame Sampling Strategy for Final QA.}
While the iterative retrieval process relies solely on textual context, visual frames are exclusively inputted to the solver during the final QA stage. To supply visual evidence for this final inference, we employ an adaptive frame sampling strategy that balances visual information richness with computational efficiency based on the preceding retrieval rounds. When employing the GPT-5 solver, we retrieve the top 10 most relevant 30-second video clips in each round. For Round 1, we uniformly sample 6 frames per clip (60 frames total), providing dense visual coverage for initial context gathering. In Round 2, we sample 6 frames only from the top 5 ranked clips (30 frames), as lower ranked clips are less likely to contain relevant visual information. From Round 3 onward, we sample 1 frame per clip (10 frames per round), since the accumulated textual context from prior rounds is typically sufficient and additional frames serve mainly as supplementary visual cues.

On EgoLifeQA~\cite{egolife}, the majority of questions (68.2\%) are resolved within a single search round, using only 60 frames. On average, each question requires 1.72 search rounds with 73.4 frames. For more complex queries where the model reaches the maximum of 5 rounds, up to 120 frames are used.

\section{Detailed Performance Comparisons}\label{appendix:comparison}

\subsection{Extended Analysis on LVBench} 
\begin{table}[t]
\centering
\caption{Category-wise performance breakdown of \ours{} and baselines on LVBench. Baseline results are taken from~\cite{worldmm}. Solvers are indicated per model (pre-trained models report weights only).}
\small
\renewcommand{\arraystretch}{1.2}
\resizebox{0.5\linewidth}{!}{%
\label{tab:sup_1_LV}
\begin{tabular}{lcccc}
\toprule
\textbf{Model} & Short & Med. & Long & Avg. \\
\midrule
\multicolumn{5}{l}{\textit{\textbf{MLLMs (Uniform Sampling)}}} \\
Qwen3-VL-8B~\cite{qwen3-vl}  & 48.8 & 44.4 & 53.4 & 48.3 \\
Gemini 2.5 Pro~\cite{gemini2.5}& 57.1 & 52.2 & 65.2 & 57.0 \\
GPT-5~\cite{gpt5} & 59.1 & 59.1 & 69.1 & 60.4 \\
\midrule
\multicolumn{5}{l}{\textit{\textbf{Hierarchical Memory Based}}} \\
EgoRAG~\cite{egolife} (GPT-5) & 32.4 & 32.0 & 31.9 & 32.2 \\
Ego-R1~\cite{egor1} (3B) & 32.5 & 36.5 & 37.3 & 34.1 \\
\midrule
\multicolumn{5}{l}{\textit{\textbf{Graph Memory Based}}} \\
LightRAG~\cite{lightrag} (GPT-5) & 30.2 & 28.6 & 34.3 & 30.4 \\
HippoRAG~\cite{hipporag} (GPT-5) & 54.9 & 47.5 & 62.3 & 54.0 \\
Video-RAG~\cite{video-rag} (GPT-5)  & 32.9 & 30.2 & 39.7 & 33.1 \\
HippoMM~\cite{hippomm} (GPT-5) & 40.7 & 33.3 & 35.8 & 38.2 \\
M3-Agent~\cite{m3-agent} (7B) & 53.0 & 40.7 & 48.5 & 49.3 \\
WorldMM~\cite{worldmm} (GPT-5)  & 58.3 & 65.4 & \textbf{72.1} & 61.9 \\ \midrule
\multicolumn{5}{l}{\textit{\textbf{Ours}}} \\
\rowcolor[gray]{0.95}
\ours{} (GPT-5) & \textbf{72.1} & \textbf{72.4} & 69.1 & \textbf{71.8} \\
\bottomrule
\end{tabular}%
}
\end{table}
\begin{table}[!t]
\centering
\caption{Category-wise performance breakdown of \ours{} and baselines on Video-MME(L)~\cite{videomme}. Baseline results are taken from~\cite{worldmm,egagent}. Solvers are indicated per model (pre-trained models report weights only).}
\small
\renewcommand{\arraystretch}{1.2} 
\label{tab:sup_2_MME}
\resizebox{1\linewidth}{!}{
\begin{tabular}{lccccccccccccc}
\toprule
\textbf{Model} & {ARES} & {AREC} & {ATTR} & {CNT} & {ISYN} & {OCR} & {ORES} & {OREC} & {SPER} & {SRES} & {TPER} & {TRES} & {Avg} \\ \midrule

\multicolumn{3}{l}{\textit{\textbf{MLLMs (Uniform Sampling)}}} \\ 
Qwen3-VL-8B~\cite{qwen3-vl} & 62.2 & 54.0 & 51.9 & 43.8 & 68.1 & 42.9 & 62.9 & 57.4 & 33.3 & 45.5 & 33.3 & 67.0 & 61.0 \\
Gemini 2.5 Pro~\cite{gemini2.5} & 56.9 & 47.6 & 66.7 & 41.7 & 71.8 & 57.1 & 53.3 & 40.7 & 0.0 & 72.7 & 66.7 & 48.4 & 55.7 \\
GPT-5~\cite{gpt5} & 71.1 & 69.8 & 70.4 & 47.9 & \textbf{88.3} & 57.1 & 75.8 & 74.1 & 33.3 & 72.7 & 50.0 & 75.8 & 74.3 \\ \midrule

\multicolumn{3}{l}{\textit{\textbf{Hierarchical Memory Based}}} \\ 
EgoRAG~\cite{egolife} (GPT-5) & 31.1 & 55.6 & 33.3 & 22.9 & 41.1 & 28.6 & 44.6 & 48.2 & 33.3 & 54.5 & 66.7 & 48.4 & 41.1 \\
Ego-R1~\cite{egor1} (3B) & 37.2 & 52.4 & 40.7 & 35.4 & 38.0 & 35.7 & 42.1 & 51.9 & \textbf{66.7} & 63.6 & 50.0 & 52.8 & 42.7 \\ \midrule

\multicolumn{3}{l}{\textit{\textbf{Graph Memory Based}}} \\ 
LightRAG~\cite{lightrag} (GPT-5) & 41.7 & 30.2 & 40.7 & 35.4 & 54.0 & 50.0 & 46.7 & 61.1 & 33.3 & 45.5 & 50.0 & 52.8 & 46.6 \\
HippoRAG~\cite{hipporag} (GPT-5) & 45.6 & 47.6 & 40.7 & 37.5 & 52.2 & 42.9 & 52.9 & 64.8 & \textbf{66.7} & 54.5 & 50.0 & 70.3 & 52.1 \\
Video-RAG~\cite{video-rag} (GPT-5) & 51.7 & 47.6 & 37.0 & 39.6 & 49.7 & 57.1 & 62.1 & 68.5 & \textbf{66.7} & 45.5 & 50.0 & 68.1 & 55.4 \\ 
HippoMM~\cite{hippomm} (GPT-5) & 41.1 & 42.9 & 55.6 & 35.4 & 38.7 & 35.7 & 37.9 & 53.7 & 33.3 & 54.5 & 50.0 & 47.3 & 41.6 \\
M3-Agent~\cite{m3-agent} (7B) & 52.2 & 57.1 & 59.3 & 45.8 & 51.5 & 42.9 & 54.6 & 64.8 & 33.3 & 45.5 & 50.0 & 71.4 & 55.3 \\
EGAgent~\cite{egagent} (Gemini 2.5 Pro) & - & - & - & - & - & - & - & - & - & -& -  & -  & 74.1 \\
WorldMM~\cite{worldmm} (GPT-5)  & \textbf{81.1} & 73.0 & 70.4 & \textbf{54.2} & 85.3 & 42.9 & 75.0 & 77.8 & 33.3 & 72.7 & 66.7 & \textbf{79.1} & 76.6 \\ \midrule

\multicolumn{3}{l}{\textit{\textbf{Ours}}} \\ 
\rowcolor[gray]{0.95}
\ours{} (Gemini 2.5 Pro) & 75.0 & \textbf{74.6} & \textbf{88.9} & \textbf{54.2} & 82.2 & \textbf{78.6} & 77.1 & \textbf{79.6} & 33.3 & \textbf{100} & \textbf{83.3} & 71.4 & 76.3 \\
\rowcolor[gray]{0.95}
\ours{} (GPT-5) &78.3 & 68.3 & 81.5 & \textbf{54.2} & 87.7 & 64.3 & \textbf{79.2} & 74.1 & 33.3 & 90.9 & \textbf{83.3} & 75.8 & \textbf{77.7} \\ \bottomrule
\end{tabular}%
}
\end{table}

\noindent\textbf{Clue Duration.} Each question in LVBench~\cite{lvbench} is annotated with a ``time reference'' field indicating the video segment required to answer it. We refer to the temporal span of this segment as the ``clue duration'' and categorize questions into three groups following prior work~\cite{worldmm}: Short (<30s), Medium (30s-5min), and Long (>5min). Of the 1,549 total questions, 1,534 are retained for this duration-based analysis after excluding 15 with missing or malformed annotations. Short-clue questions constitute the majority of the benchmark (59.6\%), followed by Medium (27.1\%) and Long (13.3\%). The mean clue duration increases substantially across groups: 8.0s for Short, 1m 33s for Medium, and 44m 52s for Long. Notably, the longest segment in the Long group spans over 2 hours, effectively requiring global comprehension of the entire video.

\vspace{0.5mm}
\noindent\textbf{Question Type Composition.} The distribution of question types varies notably across clue-duration groups. Short-clue questions are dominated by ``Entity Recognition'' (50.8\%) and ``Event Understanding'' (36.5\%), reflecting factual, moment-level queries. In contrast, Long-clue questions exhibit a markedly higher proportion of ``Summarization'' (13.2\% vs. 0.5\% in Short). These compositional differences indicate that Long-clue questions are inherently more abstract, heavily evaluating global video comprehension rather than the retrieval of localized temporal evidence.

\noindent\textbf{Accuracy by Clue Duration.} Table~\ref{tab:sup_1_LV} compares our method against previous baselines across the clue-duration groups. \ours{} achieves substantial gains on Short (+13.8\%) and Medium (+7.0\%) clue-duration questions, yielding an overall average improvement of +9.3\% points. These results demonstrate that our multi-key retrieval framework is effective for questions with localized answers, where diverse keys align with any given query to retrieve the essential context.

For the Long-clue subset, where the required context averages nearly 45 minutes and often demands full-video summarization, the performance naturally relies less on pinpoint retrieval. Nevertheless, \ours{} achieves a highly competitive score of 69.1\% on this subset, performing on par with the GPT-5 baseline that relies on dense uniform frame sampling. Fundamentally, our inference-time temporal expansion mechanism directly drives this competitive performance on summarization-heavy tasks. By dynamically expanding the retrieved episodic clips on demand, the framework provides the solver with sufficient surrounding context to synthesize higher-level semantic relations. This highlights the complementary strengths of our approach. Multi-key retrieval excels at robustly matching diverse queries to localized segments, while temporal expansion seamlessly handles queries requiring global comprehension.

\subsection{Extended Analysis on Video-MME(Long)}

\noindent\textbf{Dataset Scope and Taxonomy.} 
The Video-MME~\cite{videomme} benchmark is designed to evaluate Multimodal Large Language Models (MLLMs) on diverse video understanding tasks. It incorporates a wide range of data modalities and temporal durations, spanning 6 key domains and 30 sub-class video types with expert-annotated QA pairs. To provide a granular evaluation, the benchmark categorizes questions into 12 QA types: Action Reasoning (ARES), Action Recognition (AREC), Attribute Perception (ATTR), Counting Problem (CNT), Information Synopsis (ISYN), OCR Problems (OCR), Object Reasoning (ORES), Object Recognition (OREC), Spatial Perception (SPER), Spatial Reasoning (SRES), Temporal Perception (TPER), and Temporal Reasoning (TRES).

\vspace{0.5mm}
\noindent\textbf{Temporal Characteristics and Certificate Length.} 
One metric reported in Video-MME is the \textit{``certificate length''}, which analyzes the temporal difficulty of the QA pairs~\cite{videomme}. The certificate is defined as the minimum set of sub-clips that are both necessary and sufficient to verify the correct annotation. According to the original benchmark analysis, the ``Long'' subset features an average certificate length of approximately 16.1 minutes against an average total video length of 39.8 minutes. This indicates that answering a typical query requires comprehending roughly 40.6\% of the entire video context. Thus, this subset generally evaluates the model's ability to aggregate broad temporal information.

\vspace{0.5mm}
\noindent\textbf{Detailed Performance Breakdown.} 
Table~\ref{tab:sup_2_MME} reports the detailed performance breakdown across the 12 QA types (EGAgent is excluded as it only reports the overall average). Given the substantial 40.6\% certificate requirement, standard MLLMs employing uniform sampling establish strong baselines, suggesting that external memory architectures might be less critical for this specific dataset. This is evident in Information Synopsis (ISYN), a major QA type focusing on overall context (e.g., ``What is the main idea of the video?''), where a standard GPT-5 with uniformly sampled frames achieves the highest score. 

Despite these dataset characteristics inherently favoring standard MLLM baselines, \ours{} establishes state-of-the-art results across the majority of QA types. The key to answering such global-context queries lies in our neighbor filtering mechanism during inference-time temporal expansion. By expanding around the retrieved clips, this filtering step selectively extracts query-relevant information from the temporal neighborhood, effectively reconstructing the broader semantic context. This ensures that the solver grasps the continuous semantic flow required for holistic questions.

Looking forward, as the field advances toward true ultra-long video settings, the proportion of query-irrelevant content will drastically increase, causing the relative certificate length ratio to drop significantly. As the required context becomes a progressively smaller fraction of the entire video, tasks are expected to increasingly shift from summarizing the entire video to retrieving specific, isolated information from extensive continuous streams. Consequently, our retrieval-centric framework is highly aligned with the anticipated demands of future ultra-long video understanding.


\section{Additional Ablations and Analysis}\label{appendix:ablation}
\subsection{Round Distribution and Retrieval Hit Rate Analysis}
\begin{table}[t]
\centering
\caption{Round distribution, accuracy, and cumulative retrieval hit rate on EgoLifeQA~\cite{egolife}. ``Prop.'' denotes the proportion of questions terminated at each round. ``C.\ HR'' is the cumulative hit rate (\%) up to that round. Count, Prop., and Acc. are measured among questions terminated at each round, while Cum. HR is the cumulative proportion of all questions with at least one retrieval hit up to that round. WorldMM$^*$ denotes our reproduction of the original WorldMM~\cite{worldmm}.}
\label{tab:sup_3_round}

\resizebox{0.8\columnwidth}{!}{%
\begin{tabular}{c @{\hspace{8pt}} S[table-format=3.0] S[table-format=3.1] S[table-format=3.1] S[table-format=2.1] @{\hspace{12pt}} S[table-format=3.0] S[table-format=3.1] S[table-format=3.1] S[table-format=2.1]}
\toprule
& \multicolumn{4}{c}{\textbf{WorldMM$^*$}} & \multicolumn{4}{c}{\textbf{MERIT}} \\
\cmidrule(lr{10pt}){2-5} \cmidrule(lr{2pt}){6-9}
\textbf{Round} & {Count} & {Prop.} & {Acc.} & {C.\ HR} & {Count} & {Prop.} & {Acc.} & {C.\ HR} \\
\midrule
0 & 3 & 0.6 & 100.0 & 0.0 & 1 & 0.2 & 100.0 & 0.0 \\
1 & 214 & 42.8 & 73.4 & 37.4 & 341 & 68.2 & 74.5 & 46.4 \\
2 & 80 & 16.0 & 56.3 & 45.6 & 64 & 12.8 & 71.9 & 52.6 \\
3 & 68 & 13.6 & 51.5 & 49.2 & 32 & 6.4 & 62.5 & 54.4 \\
4 & 27 & 5.4 & 55.6 & 51.6 & 16 & 3.2 & 62.5 & 54.6 \\
5 & 108 & 21.6 & 56.5 & 53.0 & 46 & 9.2 & 54.3 & 55.6 \\
\midrule
\textbf{Total} & 500 & 100.0 & 63.2 & 53.0 & 500 & 100.0 & \textbf{71.2} & \textbf{55.6} \\
\midrule
\textbf{Avg. Rounds} & \multicolumn{4}{c}{2.45} & \multicolumn{4}{c}{\textbf{1.72}} \\
\textbf{$\le$2 Rounds} & \multicolumn{4}{c}{59.4\%} & \multicolumn{4}{c}{\textbf{81.2\%}} \\
\bottomrule
\end{tabular}%
} 
\end{table}

Following the overall retrieval evaluation presented in Table~\ref{tab:table_5_hit} of the main paper, this section analyzes the round-wise distribution and hit rates of the multi-turn QA process. We compare \ours{} against the reproduced WorldMM~\cite{worldmm} baseline, as both frameworks employ a multi-turn, agentic retrieval flow. The models are evaluated on EgoLifeQA~\cite{egolife} using GPT-5~\cite{gpt5} as the solver. Table~\ref{tab:sup_3_round} presents the detailed round-by-round breakdown. Round~0 indicates instances where the model answers immediately without any retrieval.

A key advantage of \ours{} is its high hit rate in the initial retrieval rounds. By Round~1, \ours{} achieves a cumulative hit rate of 46.4\%, outperforming WorldMM by +9.0 percentage points. Because the target evidence is effectively retrieved earlier, \ours{} can answer questions in fewer rounds. Specifically, \ours{} resolves 68.2\% of the questions within a single round at an accuracy of 74.5\%, whereas WorldMM resolves only 42.8\%. This trend continues into Round~2, where 81.2\% of \ours{}'s questions are answered (vs.\ 59.4\% for WorldMM), and its cumulative hit rate reaches 52.6\% (vs.\ 45.6\%).

Consequently, this early-round effectiveness significantly reduces the total number of retrieval iterations. \ours{} averages only 1.72 rounds compared to 2.45 for WorldMM, resulting in fewer LLM calls and faster inference. Despite this reduction in computational steps, \ours{} consistently yields higher per-round accuracy across Rounds~1 to 4 (e.g., 71.9\% vs.\ 56.3\% in Round~2). Furthermore, only 9.2\% of queries reach the maximum limit of 5 rounds in \ours{}, compared to 21.6\% in WorldMM, yet \ours{} still attains a higher final cumulative hit rate. These results demonstrate that our multi-key based retrieval, combined with contextual neighbor filtering, successfully isolates relevant segments upfront. Instead of relying on extensive multi-turn iterations to compensate for initial retrieval failures, \ours{} leverages highly effective early retrieval to minimize computational overhead while achieving a higher overall accuracy.

\begin{table}[t]
  \centering
  \begin{tabular}{@{}c@{\hspace{0.03\columnwidth}}c@{}}
    \begin{minipage}[t]{0.48\columnwidth}
      \vspace{0pt}\centering
      \captionsetup{skip=2pt}
      \caption{Comparing \ours{} against a flat retrieval baseline on the EgoLifeQA~\cite{egolife}. `Key' denotes the use of multi-key representations, and `NF' denotes neighbor filtering.}
      \label{tab:sup_4_flat}
      \resizebox{\linewidth}{!}{%
        \setlength{\tabcolsep}{4pt}
\renewcommand{\arraystretch}{1.2}
\small
\begin{tabular}{cc|rrrrrr}
  \toprule
  Key & NF & Ent. & Evt. & Hab. & Rel. & Task. & Avg. \\
  \midrule
  \ding{55} & \ding{55} & 40.0 & 48.4 & 52.5 & 43.2 & 57.1 & 46.6 \\
  \checkmark & \ding{55} & 41.6 & \textbf{54.0} & 50.8 & 42.4 & 57.1 & 48.0 \\
  \ding{55} & \checkmark & \textbf{46.4} & 50.8 & 52.5 & \textbf{53.6} & \textbf{68.3} & 52.8 \\
  \rowcolor{gray!15}\checkmark & \checkmark & 43.2 & \textbf{54.0} & \textbf{67.2} & \textbf{53.6} & 65.1 & \textbf{54.2} \\
  \bottomrule
\end{tabular}

      }
    \end{minipage}
    &
    \begin{minipage}[t]{0.48\columnwidth}
      \vspace{0pt}\centering
      \captionsetup{skip=2pt}
      \caption{Ablation on neighbor filtering window size on EgoLifeQA~\cite{egolife}. Performance across varying numbers of adjacent clips. The default ($\Delta=2$) is highlighted.}
      \label{tab:sup_5_delta}
      \resizebox{\linewidth}{!}{%
        \setlength{\tabcolsep}{4pt}
\renewcommand{\arraystretch}{1.2}
\small
\begin{tabular}{c|cccccc|c}
  \toprule
  $\Delta$ & Ent. & Evt. & Hab. & Rel. & Task. & Avg. & Hit. \\
  \midrule
  0 & 68.0 & 69.8 & 68.9 & 73.6 & 68.3 & 70.0 & 0.40 \\
  1 & 64.8 & \textbf{75.4} & 70.5 & 72.8 & 68.6 & 70.6 & 0.52 \\
  \cellcolor{gray!15}2 & \cellcolor{gray!15}67.2 & \cellcolor{gray!15}70.6 & \cellcolor{gray!15}\textbf{73.8} & \cellcolor{gray!15}74.4 & \cellcolor{gray!15}71.4 & \cellcolor{gray!15}71.2 & \cellcolor{gray!15}0.56 \\
  3 & 64.0 & 71.4 & 72.1 & \textbf{79.2} & 71.4 & 71.6 & 0.60 \\
  4 & \textbf{72.0} & 69.8 & 70.5 & 74.4 & \textbf{74.6} & \textbf{72.2} & \textbf{0.61} \\
  \bottomrule
\end{tabular}
      }
    \end{minipage}
  \end{tabular}
\end{table}

\subsection{Comparison with Flat Retrieval Baseline}

In Table~\ref{tab:sup_4_flat}, we present an ablation study on EgoLifeQA~\cite{egolife} using Qwen3-VL-8B~\cite{qwen3-vl} to compare the core components of \ours{} against a flat retrieval baseline. The flat baseline (Key={\ding{55}}, NF={\ding{55}}) performs retrieval using only standard 30-second dense caption embeddings, achieving an average score of 46.6.

By progressively integrating our proposed multi-key representations (Key) and the neighbor filtering (NF) mechanism, the overall average performance is significantly boosted to 54.2. Notably, the full configuration reaches peak scores in specific sub-categories such as EventRecall (54.0), HabitInsight (67.2), and RelationMap (53.6). These results demonstrate that both components are essential to the effectiveness of the retrieval design in \ours{}, providing a substantial improvement over a standard flat retrieval approach.

\subsection{Ablation on Neighbor Filtering Window Size}

During neighbor filtering, temporal context is constructed by including a specific number of adjacent clips, defined as the window size ($\Delta$). Table~\ref{tab:sup_5_delta} presents an ablation study on this parameter, evaluated on EgoLifeQA~\cite{egolife} using GPT-5~\cite{gpt5} with the base clip length fixed at 30-second.

Initially, expanding the temporal window yields substantial improvements in both retrieval hit rate and overall accuracy. Compared to the baseline without neighbor filtering ($\Delta$$\hphantom{}=0$), which yields a hit rate of 0.40 and an average accuracy of 70.0, integrating adjacent clips up to $\Delta$$\hphantom{}=2$ noticeably boosts the hit rate to 0.56 and the average accuracy to 71.2. At this setting, \ours{} also achieves a peak score in the HabitInsight category (73.8).

While further increasing $\Delta$ to 3 and 4 continues to incrementally improve the hit rate (up to 0.61) and average accuracy (up to 72.2), the overall rate of performance gain begins to plateau. More importantly, incorporating additional adjacent clips linearly scales the number of input tokens, which significantly increases the computational overhead during the inference stage. Therefore, to strike an optimal balance between downstream task performance and token efficiency, we establish $\Delta$$\hphantom{}=2$ as the default configuration for \ours{}.

\begin{figure}[h]
    \centering
    \includegraphics[width=0.9\linewidth]{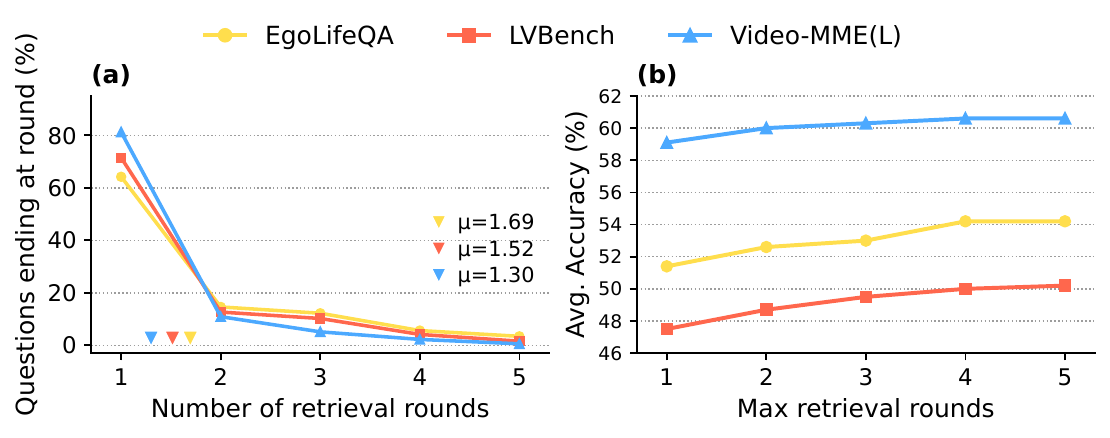} 
    \vspace{-2mm} 
    \caption{
        Retrieval round distribution and performance scaling in \ours{}. (a) The distribution of required retrieval rounds to answer a query across different benchmarks. (b) Average accuracy scaling against the maximum round budget.
    }
    \label{fig:sup_round}
\end{figure}
\vspace{-3em}

\subsection{Analysis of Retrieval Rounds}
We present a detailed analysis of the multi-turn retrieval mechanism in \ours{}, evaluated with Qwen3-VL-8B~\cite{qwen3-vl} across multiple long video benchmarks. Fig.~\ref{fig:sup_round}~(a) illustrates the distribution of required retrieval rounds to answer a query, capped at a default maximum of 5 rounds. The results show that most queries are resolved highly efficiently in the early stages, with average retrieval rounds ($\mu$) of 1.69 for EgoLifeQA~\cite{egolife}, 1.52 for LVBench~\cite{lvbench}, and 1.30 for Video-MME(L)~\cite{videomme}. This empirical evidence demonstrates a direct correlation between the video length and the required retrieval budget; longer video contexts naturally necessitate slightly more iterative searches.

Additionally, Fig.~\ref{fig:sup_round}~(b) plots the performance scaling against the maximum round budget across all three benchmarks. As the round limit increases, the average accuracy steadily improves and eventually converges. This consistent trend confirms that the multi-turn mechanism in \ours{} effectively resolves complex questions requiring iterative refinement, while maintaining computational efficiency for simpler queries.


\section{Discussion}\label{appendix:discussion}
\subsection{Solver Dependency and Intelligence Allocation}
\noindent\textbf{Perspective on Memory Building.} 
From the perspective of memory building, \ours{} demonstrates effective retrieval capabilities. As evidenced by the hit rate results, the lightweight episodic memory construction successfully locates relevant temporal evidence without relying on complex, pre-computed hierarchical graph structures. This suggests that elaborate structural pre-computation of the memory is largely redundant, as the inherent richness of the raw episodic memory itself provides sufficient grounding for accurate retrieval.

\noindent\textbf{Perspective on Solver and Intelligence Allocation.}
From the solver's perspective, however, post-retrieval reasoning emerges as the critical factor for overall performance. As shown in Table~\ref{tab:table_1_EgoLifeQA} of the main paper, upgrading the solver from Qwen3-VL-8B~\cite{qwen3-vl} to GPT-5~\cite{gpt5} yields a substantial 17\% performance improvement. Crucially, the underlying episodic memory $M$ remains identical across both settings. This confirms that once the relevant evidence is retrieved, the solver's inherent capacity to process the provided information plays a significantly more decisive role than the structural complexity of the memory itself.

Furthermore, this insight highlights a critical inefficiency in conventional memory-based approaches. Prior methods typically require computationally expensive MLLMs for both pre-computing complex hierarchical graphs and answering the final queries, resulting in redundant intelligence usage. Consequently, our empirical findings and the structural observations of previous baselines strongly support our core motivation of \textit{intelligence allocation}. Rather than engineering complex static memory structures upfront with redundant computational costs, we shift the primary intelligence requirement entirely to the inference stage. By providing rich context through temporal expansion, we enable capable solvers to maximize their reasoning potential.

\subsection{Future Works}

By allocating the high-level reasoning to capable MLLMs, \ours{} maintains a strictly lightweight episodic memory. Without the structural overhead of constantly updating complex memory, our framework is naturally suited for lifelong video understanding. Future work will explore scaling this multi-key retrieval approach to effectively process unbounded, continuous video streams.

\section{Additional Qualitative Analysis}\label{appendix:qualitative}
\subsection{Qualitative Comparison with Memory-based Baselines}

In this section, we present a qualitative comparison of \ours{}, against two memory-based baselines: EgoRAG~\cite{egolife} (hierarchy-based memory) and WorldMM\cite{worldmm}(graph-based memory), evaluated on the EgoLifeQA~\cite{egolife}. EgoRAG constructs a temporal hierarchy (ranging from 30-second clips to day-level granularity) and retrieves evidence by matching question keywords with text embeddings within a heuristically determined time scope. WorldMM constructs three types of multimodal memory: episodic, semantic, and visual. When querying its episodic memory, WorldMM searches a multi-granularity episodic graph and employs multi-round query reformulation, retrieving relevant nodes based on Personalized PageRank (PPR) scores. 

\begin{figure}[t]
  \centering
  \setlength{\abovecaptionskip}{4pt}
  \setlength{\belowcaptionskip}{2pt}
  \includegraphics[width=1\linewidth]{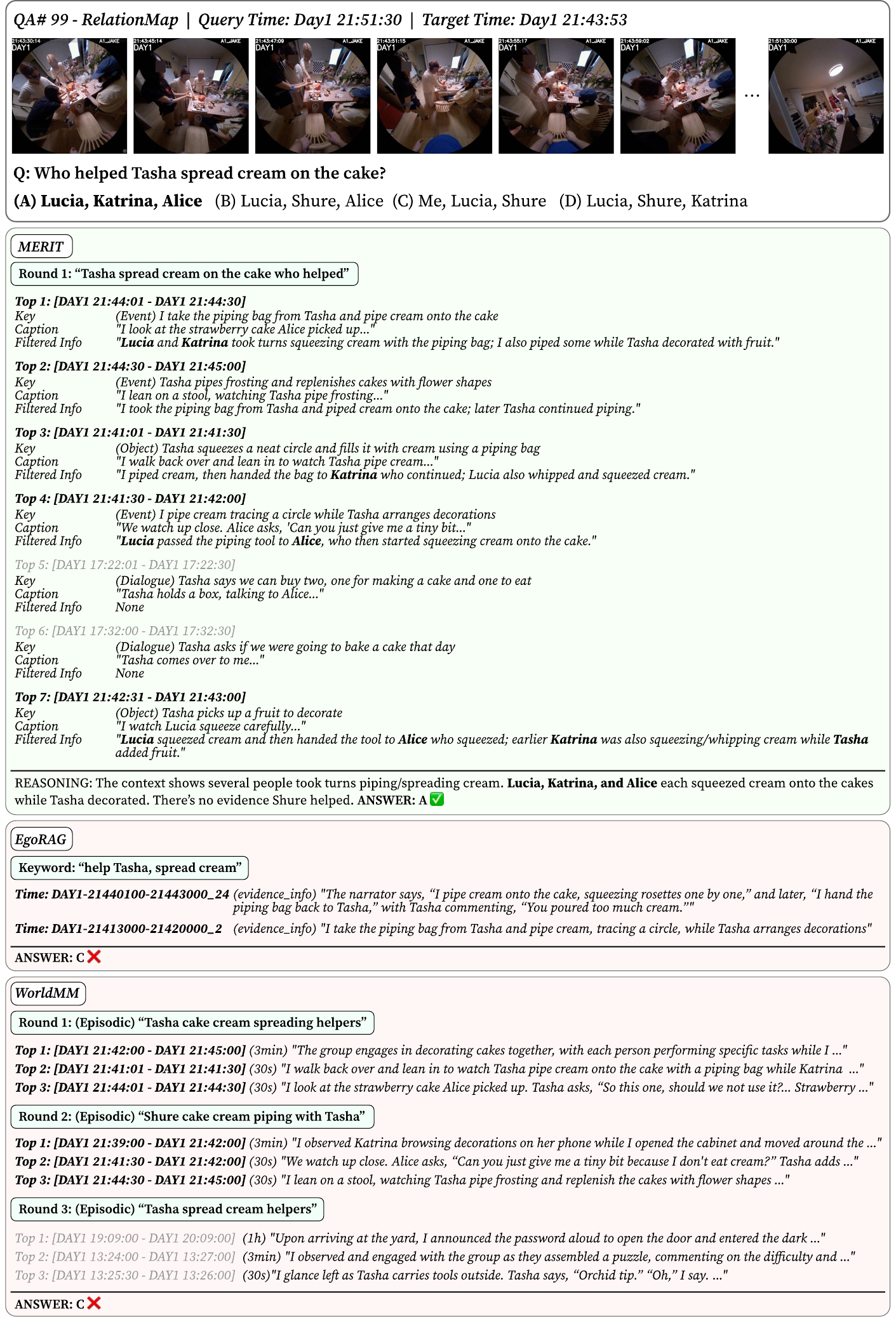}
  \caption{Qualitative comparison on a \textit{RelationMap} query, where \ours{} successfully resolves complex interpersonal interactions using Neighbor Filtering, while other baselines fail.}
  \label{fig:placeholder-sup1}
\end{figure}

Fig.~\ref{fig:placeholder-sup1} illustrates a \textit{RelationMap} query with a short temporal context (query time at Day 1). The question, \textit{``who helped Tasha spread cream on the cake?''}, requires not only temporal localization but also precise relationship mapping among multiple individuals in the scene. While all three models localize the target time window, EgoRAG and WorldMM fail to identify the helpers, erroneously including the camera wearer (``I''). In contrast, \ours{} accurately resolves these complex interpersonal interactions. Our Neighbor Filtering module effectively synthesizes the surrounding context to pinpoint the exact individuals and their corresponding actions, providing accurate filtered information.

\begin{figure}[t]
  \centering
  \setlength{\abovecaptionskip}{4pt}
  \setlength{\belowcaptionskip}{2pt}
  \includegraphics[width=1\linewidth]{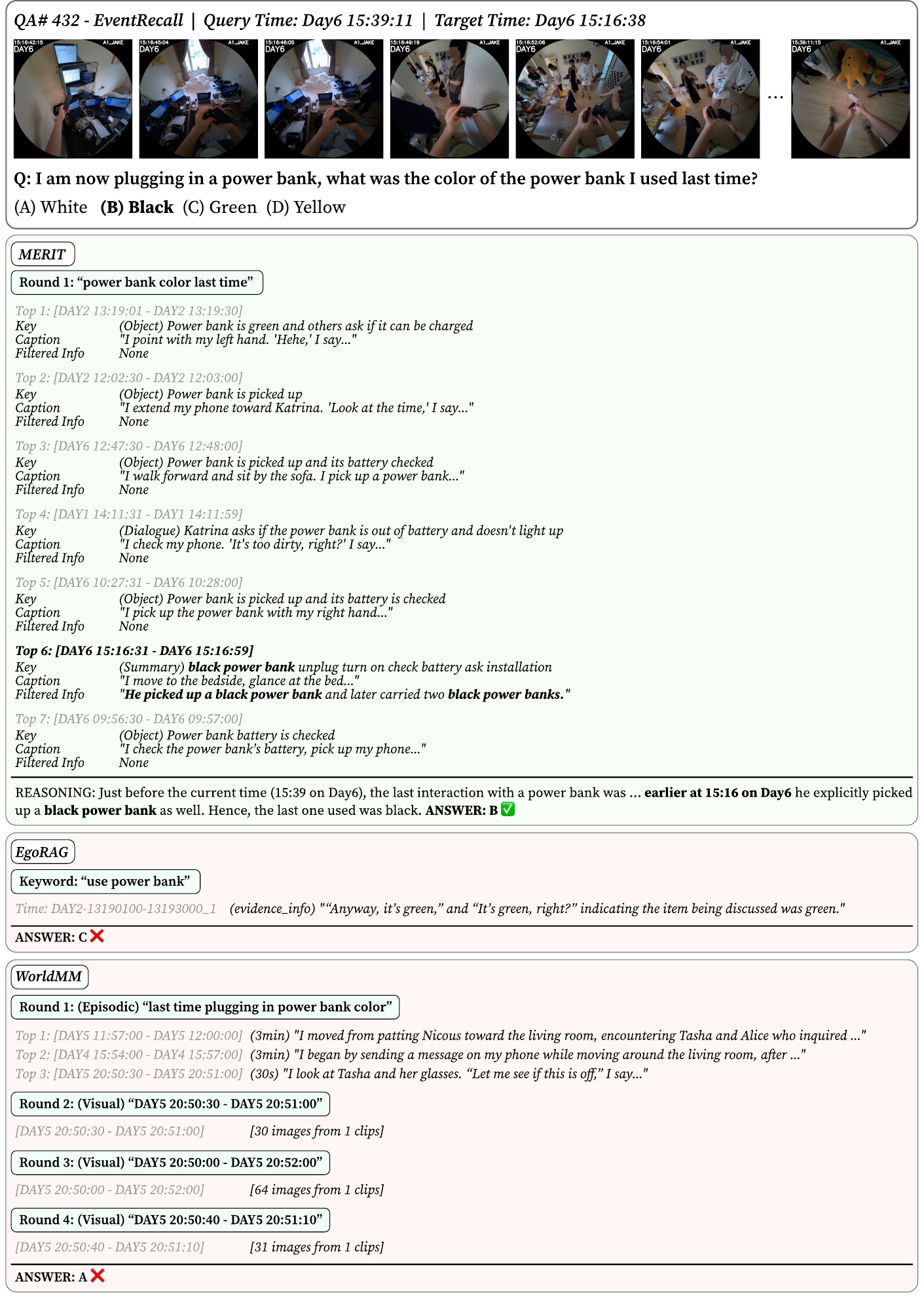}
  \caption{Qualitative comparison on an \textit{EventRecall} query over an extended temporal horizon, demonstrating that \ours{} accurately localizes fine-grained object details using the pre-extracted Summary Key while baselines fail to find the target time.}
  \label{fig:placeholder-sup2}
\end{figure}

Fig.~\ref{fig:placeholder-sup2} demonstrates an \textit{EventRecall} query over an extended temporal horizon (query time at Day 6). The query demands fine-grained object details \textit{(e.g., the color of a previously used power bank)}. EgoRAG fails to localize the correct target time entirely, and WorldMM also fails despite its multi-round retrieval attempts. \ours{}, however, successfully retrieves the target clip because the pre-extracted Summary Key explicitly preserves critical fine-grained attributes \textit{(e.g., ``black power bank'')}, enabling direct and accurate localization.

\begin{figure}[t]
  \centering
  \setlength{\abovecaptionskip}{4pt}
  \setlength{\belowcaptionskip}{2pt}
  \includegraphics[width=1\linewidth]{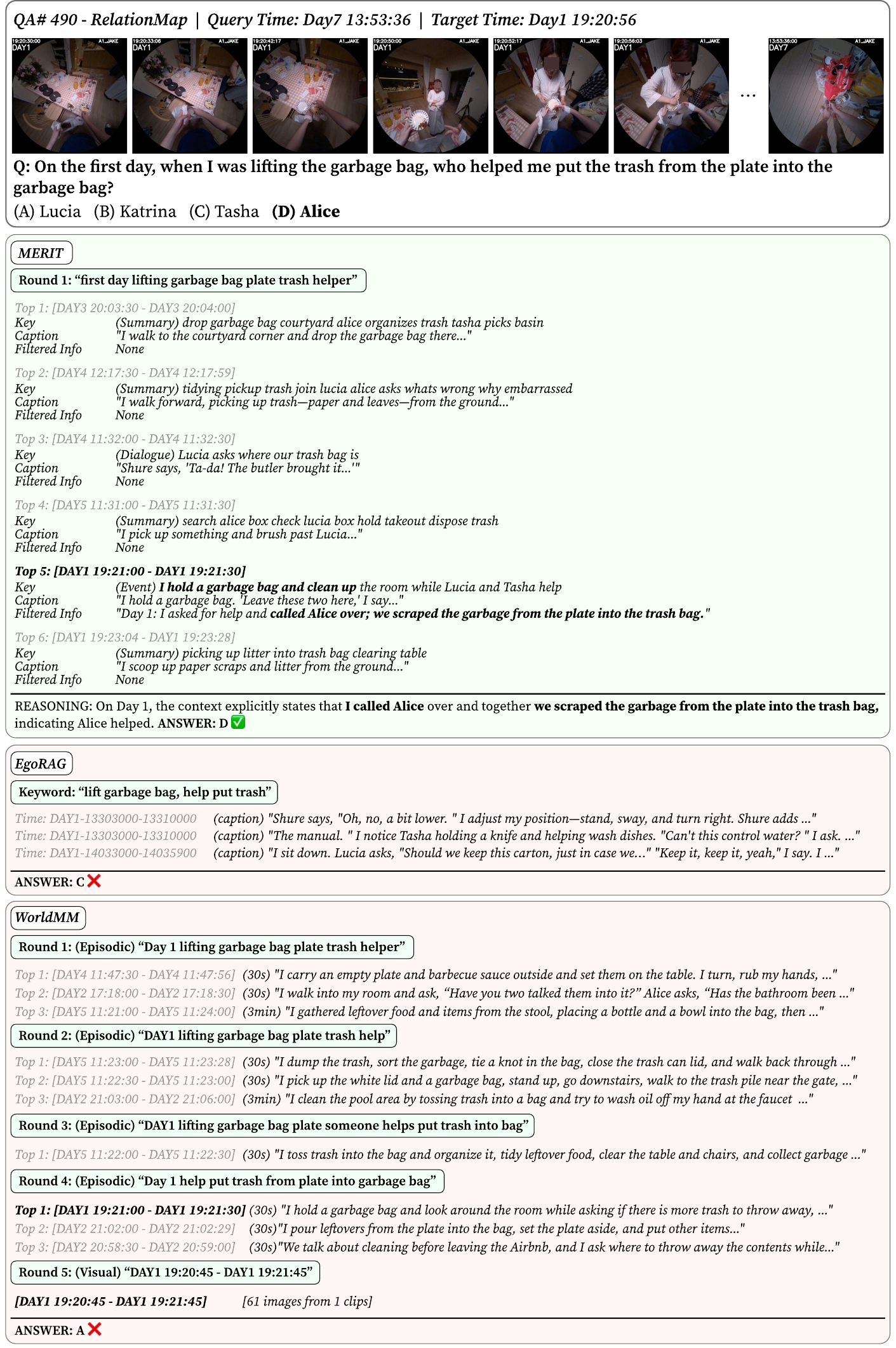}
  \caption{Qualitative results for a query with an extreme temporal gap, highlighting that \ours{} efficiently retrieves the correct context in a single round via the Event Key.}
  \label{fig:placeholder-sup3}
\end{figure}

Fig.~\ref{fig:placeholder-sup3} highlights a scenario with an extreme temporal gap between the target event (Day 1) and the query time (Day 7). \ours{} achieves accurate retrieval in a single round, as the pre-extracted Event Key robustly captures the specific objects and actions requested by the query. Subsequently, the Neighbor Filtering module expands the local temporal context to extract the precise evidence required to answer the question \textit{(e.g., asking Alice for help)}. Conversely, EgoRAG completely fails to localize the target time. While WorldMM eventually locates the target time after five retrieval rounds, it still yields an incorrect answer due to a lack of sufficient contextual information in its retrieved memory.

\subsection{Failure Cases}
\begin{figure}[t]
  \centering
  \setlength{\abovecaptionskip}{4pt}
  \setlength{\belowcaptionskip}{2pt}
  \includegraphics[width=1\linewidth]{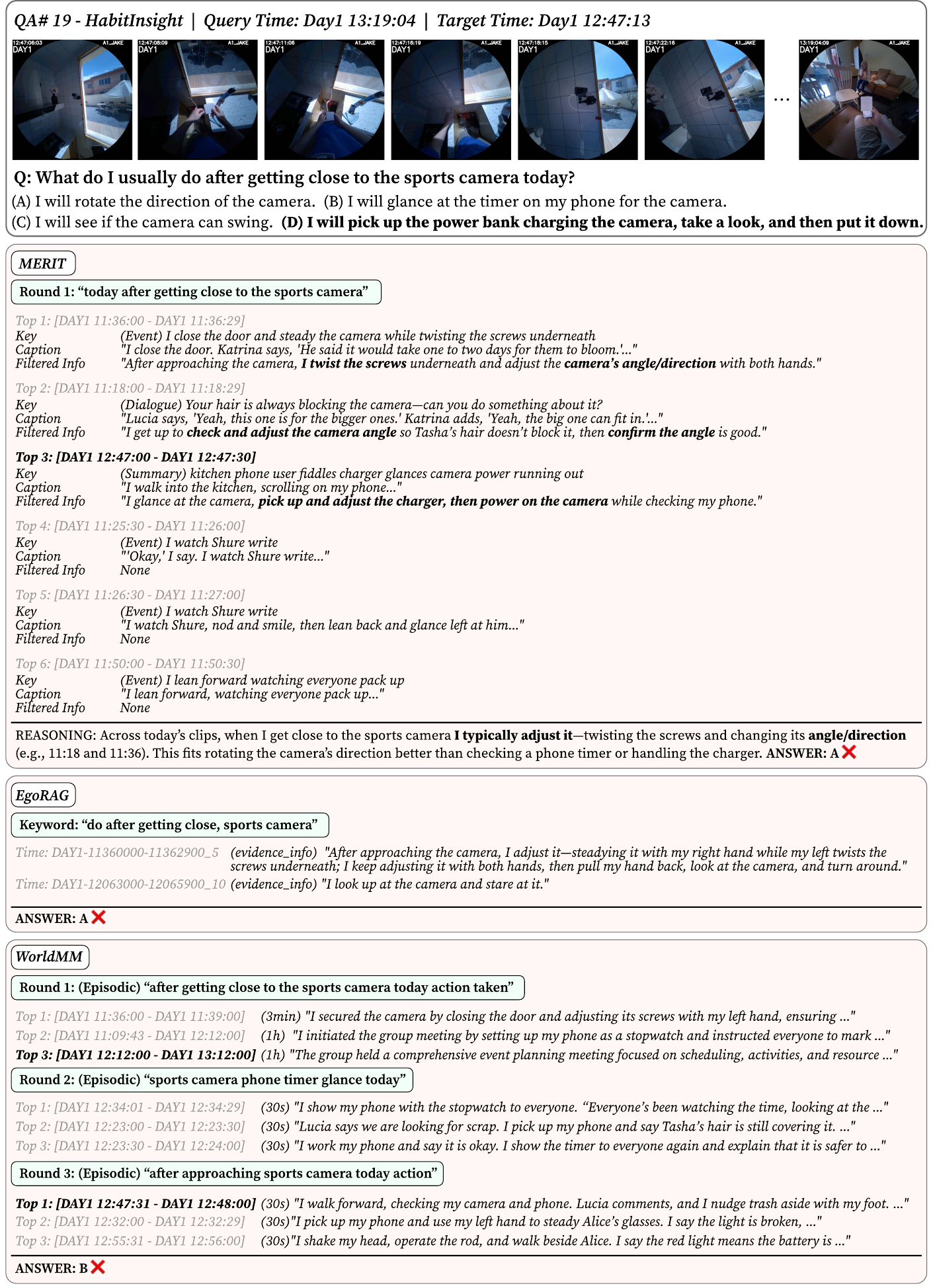}
  \caption{A failure case involving a frequency query, where the QA solver is misled by incorrect actions in the top-ranked clips.}
  \label{fig:placeholder-sup4}
\end{figure}

Fig.~\ref{fig:placeholder-sup4} illustrates a typical failure case involving frequency or counting queries \textit{(e.g., ``how many times'' or ``usually'')}. A fundamental limitation of current retrieval-based systems is that they return a limited set of individual events, rather than aggregating all relevant instances across the entire video timeline. While \ours{} successfully retrieves the correct evidence within its top-3 results, the QA solver is ultimately misled because the higher-ranked clips (top-1 and top-2) support the incorrect Answer A. Consequently, queries that demand global temporal aggregation or routine reasoning remain a challenging direction for memory-based video understanding.

\clearpage

\section{Prompt Appendix}\label{appendix:prompts}
\label{sec:prompts}


\definecolor{maincolor}{RGB}{119, 178, 255}
\definecolor{backcolor}{RGB}{248, 250, 252}

\begin{center}
\begin{tcolorbox}[
    enhanced jigsaw,
    breakable,
    colback=backcolor,
    colframe=maincolor,
    boxrule=0.7pt,
    title=\textbf{Multi-key Extraction Prompt},
    fonttitle=\bfseries, 
    colbacktitle=maincolor,
    top=1mm,    
    bottom=1mm,
    left=4mm,
    right=4mm,
    sharp corners,
    drop shadow=gray!30
]
\begin{lstlisting}[
    basicstyle=\fontsize{7pt}{8.2pt}\ttfamily, 
    aboveskip=0pt, 
    belowskip=0pt,
    breaklines=true,
    breakatwhitespace=true,
    columns=fullflexible,
    keepspaces=true,
    showstringspaces=false
]
You are extracting retrieval keys from an episodic video memory clip.

Each input value corresponds to a ~30-second video clip and consists of:
- physical actions and movements
- spoken dialogue between people
- interactions with objects
- reflect the entire clip by summarizing the clip

Your task is to extract EXACTLY FOUR retrieval keys from the value.
Do NOT write extra explanations.
Do NOT invent events.
Use only information explicitly present in the value.

The four keys MUST correspond to the following categories:

1. Event / Action key
- What physical actions or events actually happened between people?
- Focus on observable actions and interactions.
- Use one short sentence or phrase.
- Include the agent if identifiable (use actual names if present).
- If the speaker uses first-person expressions (I / me), use 'I' or 'me'.

2. Dialogue / Mention key
- What was said, asked, or mentioned in the dialogue?
- Focus on questions, statements, commands, or repeated mentions.
- Use one short sentence or phrase.

3. Object-state / Item-centric key
- What object was handled, requested, moved, or referenced?
- Describe the object and its state or role in the scene.
- Use one short sentence or phrase.

4. Summary / Retrieval key
- Generate ONE concise retrieval key that best represents the core event of the clip.
- Abstract away redundant or repeated actions.
- Capture the main entities, actions, and intent.
- Be concise and retrieval-friendly.
- Prefer compact keyword-style phrasing (not a full sentence).
- Use spaces between words.
- Stay grounded in the value; do not add details.

Formatting rules:
- Output exactly four lines
- One key per line, in the order: event, dialogue, object, summary
- Use spaces between words
- DO NOT use underscores (_)
- Do not include numbering, bullets, or extra explanations
- Each of lines 1-3 must be a single sentence or a single clause
- Line 4 should be a short keyword-style phrase (not necessarily a sentence)

# Few-shot Examples: [... Few-shot examples ...]

Now extract the four retrieval keys from the following value.

Value:
{caption}
\end{lstlisting}
\end{tcolorbox}
\captionof{figure}{Prompt used for Multi-key Extraction.} 
\label{fig:multi_key_prompt} 
\end{center}


\definecolor{pinkmain}{RGB}{255, 168, 168} 
\definecolor{pinkback}{RGB}{253, 248, 250} 

\begin{center}
\begin{tcolorbox}[
    enhanced jigsaw,
    breakable,
    colback=pinkback,
    colframe=pinkmain,
    boxrule=0.7pt,
    title=\textbf{Neighbor Filtering Prompt}, 
    fonttitle=\bfseries, 
    colbacktitle=pinkmain,
    top=1mm,    
    bottom=1mm,
    left=4mm,
    right=4mm,
    sharp corners,
    drop shadow=gray!30
]
\begin{lstlisting}[
    basicstyle=\fontsize{7pt}{8.2pt}\ttfamily, 
    aboveskip=0pt,
    belowskip=0pt,
    breaklines=true,
    breakatwhitespace=true,
    columns=fullflexible,
    keepspaces=true,
    showstringspaces=false,
    literate={±}{{$\pm$}}1
]
You are a helpful assistant that extracts relevant information from video captions.

Given a question with multiple choice answers and captions from retrieved clips' neighborhoods (±1 minute window each), your task is to:
1. For each retrieved clip, analyze its 5 neighbor captions (before_2, before_1, center, after_1, after_2)
2. Extract ONLY the information relevant to answering the question
3. Return relevant info for each clip

Output format (JSON):
{
  "clip_1": "...concise relevant info...",
  "clip_2": "",
  ...
  "clip_N": "..."
}

Guidelines:
- Focus on information that directly helps answer the question
- If no relevant information is found for a clip, output empty string ""
- Keep each relevant_info concise (1-3 sentences)
- Output valid JSON only, no extra commentary
\end{lstlisting}
\end{tcolorbox}
\captionof{figure}{Prompt used for Neighbor Filtering.}
\label{fig:neighbor_filtering_prompt}
\end{center}

\end{document}